\documentclass[manuscript,screen]{acmart}

\usepackage{url,hyperref,lineno,microtype,subcaption}
\usepackage{comment}
\usepackage{adjustbox}
\usepackage{makecell}
\usepackage{float}
\usepackage{pdflscape}
\usepackage{afterpage}
\usepackage{bigstrut}
\usepackage{booktabs,rotating,caption}
\usepackage{graphicx}
\usepackage{caption}
\usepackage{array}
\usepackage{booktabs}
\usepackage{multicol}
\usepackage{longtable,tabularx,ltxtable}
\usepackage{ragged2e}  
\usepackage{natbib}
\usepackage{amsthm}
\usepackage[ruled,vlined]{algorithm2e}
\newcolumntype{P}[1]{>{\RaggedRight\hspace{0pt}}p{#1}}
\newcolumntype{Y}[1]{>{\RaggedRight\arraybackslash}p{#1}}
\newcolumntype{L}{>{\RaggedRight\arraybackslash}X} 
\newcolumntype{C}{>{\Centering\arraybackslash}X}   

\newcommand{\qt}[1]{``#1''}
\newcommand{\qo}[1]{`#1'}
\DeclareMathOperator*{\argmax}{argmax}

\begin{document}
\onecolumn

\title[Reinforcement Learning Approaches in Social Robotics]{Reinforcement Learning Approaches in Social Robotics} 

\author{Neziha Akalin}
\email{neziha.akalin@oru.se}
\affiliation{
  \institution{\"{O}rebro University}
  \streetaddress{School of Science and Technology}
  \city{\"{O}rebro}
  \postcode{SE-701 82}
  \country{Sweden}
  }

\author{Amy Loutfi}
\email{amy.loutfi@oru.se}
\affiliation{%
  \institution{\"{O}rebro University}
  \streetaddress{School of Science and Technology}
  \city{\"{O}rebro}
  \postcode{SE-701 82}
  \country{Sweden}
  }

\begin{abstract}
This article surveys reinforcement learning approaches in social robotics. Reinforcement learning is a framework for decision-making problems in which an agent interacts through trial-and-error with its environment to discover an optimal behavior. Since interaction is a key component in both reinforcement learning and social robotics, it can be a well-suited approach for real-world interactions with physically embodied social robots. The scope of the paper is focused particularly on studies that include social physical robots and real-world human-robot interactions with users. We~present a thorough analysis of reinforcement learning approaches in social robotics. In addition to a survey, we categorize existent reinforcement learning approaches based on the used method and the design of the reward mechanisms. Moreover, since communication capability is a prominent feature of social robots, we discuss and group the papers based on the communication medium used for reward formulation. Considering the importance of designing the reward function, we~also provide a categorization of the papers based on the nature of the reward. This categorization includes three major themes: interactive reinforcement learning, intrinsically motivated methods, and task performance-driven methods. The benefits and challenges of reinforcement learning in social robotics, evaluation methods of the papers regarding whether or not they use subjective and algorithmic measures, a discussion in the view of real-world reinforcement learning challenges and proposed solutions, the points that remain to be explored, including the approaches that have thus far received less attention is also given in the paper. Thus, this paper aims to become a starting point for researchers interested in using and applying reinforcement learning methods in this particular research~field.

 
\end{abstract}
\maketitle

\section{Introduction}

With the proliferation of social robots in society, these systems will impact users in several facets of life from providing assistance, performing cooperation, and taking part in collaboration tasks. In order to facilitate natural interaction, researchers in social robotics have focused on robots that can adapt to diverse conditions and to different user needs. Recently, there has been great interest in the use of machine learning methods for adaptive social robots~\cite{Keizer2014, de2015robots, hemminghaus2017, ritschel2019adaptive}. Machine Learning (ML) algorithms can be categorized into three sub fields: supervised learning, unsupervised learning and reinforcement learning. In supervised learning, correct input/output pairs are available and the goal is to find a correct mapping from input to output space. In unsupervised learning, output data is not available and the goal is to find patterns in the input data. Reinforcement Learning (RL), on~the other hand, is a framework for decision-making problems in which an agent interacts through trial-and-error with its environment to discover an optimal behavior~\cite{sutton1998introduction}. The RL agent receives scarce feedback about the actions it has taken in the past. The~agent then tunes its behavior over time via this feedback signal, i.e., reward or penalty. The~agent's goal is therefore learning to take actions that maximize the reward. 

RL approaches are gaining increasing attention in the robotics community. As interaction is a key component in both RL and social robotics, RL could provide a suitable approach for social human-robot interaction. Worth noting is that humans perform sequential decision-making in daily life where sequential decision making describes problems that require successive observations, i.e., cannot be solved with a single action~\cite{barto1989learning}. Consequently, much of social human-robot interactions can be formulated as sequential decision-making tasks, i.e., RL problems. The goal of the robot in these types of  interactions would be to learn an action-selection strategy in order to optimize some performance metric, such as user satisfaction.

\begin{figure}[!ht]
\begin{center}
\includegraphics[width=0.85\textwidth]{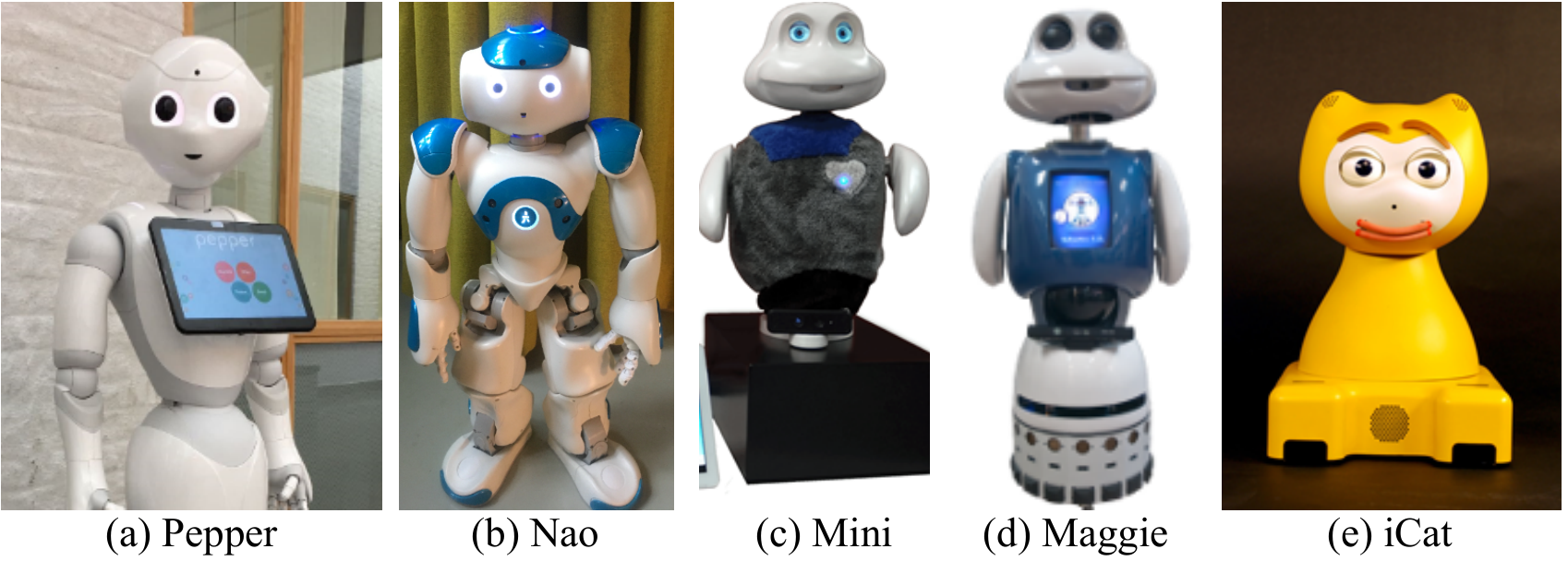}
\end{center}
\caption{Some of the social robots referenced within in the reviewed papers.}\label{fig:social_robots}
\medskip
\small
The pictures of (a) Pepper robot, and (b) Nao robot were taken by the authors. (c) Mini robot, the figure is adapted from~\citep{maroto2018bio} --- licensed under the Creative Commons Attribution, (d) Maggie robot, the figure is from~\url{https://robots.ros.org/maggie/}, accessed 20 March, 2020 --- licensed under the Creative Commons Attribution, (e) iCat robot, the figure is from~\url{https://www.bartneck.de/wp-content/uploads/2009/08/iCat02.jpg}, accessed on 22 March 2020--- used with permission, photo credit to Christoph Bartneck. 
\end{figure}

Before outlining the research related to reinforcement learning in social robots, first it is important to establish the definition of a social robot in the context of this article. A variety of definitions for a social robot have been proposed in the literature~\cite{fong2003survey, breazeal2003toward, duffy2003anthropomorphism, bartneck2004design, hegel2009understanding, yan2014survey}. Within each of these definitions, there is a wide spectrum of characteristics. However, two important aspects become prominent in these definitions that are considered in this paper, namely, embodiment and interaction/communication capability. One example can be found in~\citet{bartneck2004design} where they define a social robot as an~\qt{... autonomous or semi-autonomous robot that interacts and communicates with humans by following the behavioral norms expected by the people with whom the robot is intended to interact.} Following this definition, the authors stress that a social robot must have a physical embodiment. Based on the presented definitions in~\cite{fong2003survey, breazeal2003toward, duffy2003anthropomorphism, bartneck2004design, hegel2009understanding, yan2014survey}, we consider social robots as embodied agents that can interact and communicate with humans. Figure~\ref{fig:social_robots} shows some of the social robots that are used in the reviewed papers. 

This article presents a survey on RL approaches in social robotics. As such, it is important to emphasize that the scope of this survey is focused on studies that include physically embodied robots and real-world interactions. Considering the definition of~\cite{bartneck2004design} given above, this paper excludes studies with simulations and virtual agents where no physical embodiment is present. The presented review also excludes studies with industrial robots and studies that do not include any interaction with humans. Rather, this review exclusively focuses on papers that comprise both a social robot(s) and human input/user studies. It is worth noting that studies which use simulations for training and test on physical robot deployment with user studies fall within the selection criteria. Likewise, studies that use explicit or implicit human input in the learning process are also included. 

Due to the complexity of the social interactions and the real-world, most of the studies applying RL are trained and tested in simulation environments. However, real-world interactions are extremely important not only for social robots but also for understanding the full potential of reinforcement learning. It is mentioned in~\cite{rlbooknew} (p.~391), that \qt{the full potential of reinforcement learning requires reinforcement learning agents to be embedded into the flow of real-world experience, where they act, explore, and learn in our world, and~not just in their worlds.} Generally speaking, the overall goal of an RL agent is to maximize the expected cumulative reward over time, as stated in the~\qt{reward hypothesis}~\cite{rlbooknew} (p.~42). The reward in RL is used as a basis for discovering an optimal behavior. Hence, reward design is extremely important to elicit desired behaviors in RL-based systems. The choice of reward function is crucial in robotics, where the problem is also referred to as the~\qt{curse of goal specification}~\cite{kober2013reinforcement}. Therefore, in this paper, we provide a categorization based on reward design which is crucial for RL to be successful.
Moreover, since communication capability is a distinctive feature of social robots, we discuss communication mediums utilized for reward design together with RL algorithms. 

Finally, it is also worth noting that in the general field of robotics there is a plethora of research in RL. There also exist review papers on the topic of RL in robotics such as applications of RL in robotics in general~\cite{kober2013reinforcement, kormushev2013reinforcement}, policy search in robot learning~\cite{deisenroth2013survey}, safe~RL~\cite{garcia2015comprehensive}, and Deep Reinforcement Learning (DRL) in soft robotics~\cite{bhagat2019deep}. Indeed, RL has been applied to a variety of scenarios and domains within social robotics, with growing popularity. While~the field of social robotics deserves a survey on its own, to the best of our knowledge, there~exists no such survey on this particular research field. Thus, the main purpose of this work is to serve as a reference guide that provides a quick overview of the literature for social robotics researchers who aim to use RL in their research. Depending on the target user group, the application domain or the experimental scenario, different types of rewards, problem formulations or algorithms can be more suitable. In that sense, we believe that this survey paper will be beneficial for social robotics researchers.

\subsection*{Overview of the Survey}
After surveying research on RL and social robotics, we analyze and categorize the studies based on four different criteria: (1) RL type, (2) the utilized communication mediums for reward function formulation, (3) the nature of the reward function, (4) the evaluation methodologies of the algorithms. These categorizations aim to facilitate and guide the choice of a suitable algorithm by social robotics researchers in their application domain. For~that purpose, we elaborate on the different methods that are tested in real-world scenarios with a physical robot.

Categorization based on RL type includes bandit-based methods, value-based methods, policy-based methods, and deep RL (see Section~\ref{categories_rl_based}). The utilized communication mediums are verbal communication, nonverbal communication, affective communication, tactile communication, and additional communication medium between the robot and the human. Moreover, there are studies in which higher interaction dynamics are used for reward formulation such as engagement, comfort, and attention. There are also other studies that do not use any communication medium at all for reward formulation. In the categorization based on the design of the reward mechanisms, three major themes emerged:  
\begin{enumerate}
\item Interactive reinforcement learning: 
In these methods, humans are involved in the learning process either by providing reward or guidance to the agent (Section~\ref{interactive_rl}). This~approach, in which the human delivers explicit or implicit feedback to the agent, is known as Interactive Reinforcement Learning (IRL). 

\item Intrinsically motivated methods: There are different intrinsic motivations in the literature on RL~\cite{oudeyer2008can}, however, the most frequently used approaches in social robotics depend on the robot maintaining an optimal internal state by considering both internal and external circumstances (Section~\ref{intrinsically_motivated}). 

\item Task performance driven methods: In these methods, the reward the robot receives depends on either the robot's task performance or the human interactant's task performance, or a combination of both (Section~\ref{task_performance}). 
\end{enumerate}

The evaluation methodologies include (1) the algorithm point of view, (2) the user experience point of view, and (3) evaluation of both learning algorithm-related factors and user experience-related factors. 

To formulate the social interactions as a reinforcement learning problem, researchers need to consider some key concepts such as input data, state representation, robot actions, and reward function. Moreover, after the implementation of RL, it should be decided how the evaluation will be performed. Therefore, we extract from each of the cited works the following key points (1) the input data, state space and action space (2) the reward function (3) the communication medium in the HRI scenario (4) the main experimental results (5) the experimental scenario and its validation. Therefore, the contributions of this paper include: (i) analysing and categorising the relevant literature in terms of type of RL used; (ii) analysing and categorising the relevant literature based on the reward function; (iii)~analysing the relevant literature in terms of evaluation methodologies. 

The paper is organized as follows: In Section~\ref{benefits_and_challenges}, we discuss the benefits and challenges of applying RL in the social robotics domain. In Section~\ref{background}, we present a background on reinforcement learning. Following the formal presentation of the methods, in Section~\ref{categories_rl_based}, we present the applications of these methods in social robotics. Later, we present the categorization based on reward functions in Section~\ref{sec:categorization}. Evaluation methods are discussed in Section~\ref{evaluation_methods}. In Section~\ref{discussions}, we discuss the current approaches in the view of real-world RL challenges and proposed solutions. The section further includes the points that remain to be explored, and the approaches that have thus far received less attention. Finally, in Section~\ref{conclusions}, we conclude the paper.

\section{RL in Social Robotics---Benefits and Challenges}
\label{benefits_and_challenges}

Applications of social robots are numerous and range from entertainment to eldercare. The~robot tasks in such cases involve interactive elements such as human-robot cooperation, collaboration, and assistance. To achieve longitudinal interaction with social robots, it is important for such robots to learn incrementally from interactions, often with non-expert end-users. In~consideration of continuously evolving interactions where user needs and preferences change over time, hand-coded rules are labor-intensive. Even~though rule-based systems are deterministic, it can be difficult to create rules for complex interaction patterns. Machine learning is bound to play an important role in a wide range of domains and applications including robotics. However, the social robot learning problem differs from the traditional ML setting in which there is a need for collected datasets or assumptions about the distribution of input data~\cite{ thomaz2016computational}. Often, social robots should be able to learn new tasks and task refinements in domestic (unstructured) environments. Furthermore, social robotics researchers need to deal with a particular challenge of learning in real-time from human-robot interactions. ML paradigms such as supervised learning and unsupervised learning are not designed for learning from real-time social interactions. On the contrary, RL represents an active process. Unlike other ML methods, it does not need to be provided desired outputs instead, it trains interactively based on reward signals and refines its behavior throughout the interaction. Moreover, interaction is a key component for social robots which makes RL a suitable approach. RL also provides a possibility to learn from natural interaction patterns by utilizing the various social elements in the learning process. Consideration of all these points suggests that socially guided machine learning~\cite{thomaz2006socially} could be a more suitable approach than traditional ML approaches for social~HRI.

In general, combining human and machine intelligence may be effective for solving computationally hard problems~\cite{holzinger2016towards}. The term \qt{socially guided machine learning} was first used by~\citet{thomaz2006socially} and refers to approaches that include social interaction between a user and a machine in the learning process. Studies using IRL in social robotics can be considered as socially guided machine learning since they make use of human feedback in different forms in the learning process. The feedback provided by the human can be used for shaping the action policy (the human is involved in the action selection mechanism), or shaping the reward function~\cite{knox2012reinforcement}. It can be treated either as reward, in that the feedback is given based on the agent's past actions indicating \qt{how good the taken action was}, or policy feedback in which human feedback affects action selection or modification thereby indicating \qt{what to do}.

The majority of studies included in this review paper use IRL which may suggest that IRL could be the best suited approach in social robotics. However, IRL has its own challenges. Human~teachers tend to give less frequent feedback (due to boredom and/or fatigue) as learning progresses, resulting in diminished cumulative reward~\cite{isbell2001social}. Likewise, human~teachers tend to provide more positive reward than punishment~\cite{suay2011effect,thomaz2006reinforcement}. Yet another problem in IRL is the transparency issues that might arise during the training of a physical robot via human reward~\cite{thomaz08sociallyguided,Knox2013}. Reference \cite{Knox2013} used an audible alarm to alert the trainer about the robot's loss of sense. ~\citet{suay12} observed that experts could teach the defined task in a predefined time frame, whereas the same amount of time was not enough for inexperienced users. One solution suggested for this was algorithmic transparency during training, which shows the internal policy to the human teacher. However, the~presentation of the model of the agent's internal policy might be obscure for naive human teachers. Therefore, this information should be presented in a straight-forward way that is easy to understand to avoid causing confusion. To exemplify, in~\cite{thomaz08sociallyguided} human trainers waited for the Leonardo robot to establish eye contact with them before they continued teaching. The eye contact was considered as the robot being ready for the next action. These kinds of transparent behaviors in which the robot communicates the internal state of the learning process should be taken into account for guiding human trainers in IRL. As noted in several studies, in IRL, the human teacher's positive and negative reward can be much more deliberate than a simple \qo{good} or \qo{bad} feedback~\cite{Thomaz2007, thomaz08sociallyguided}. The learning agent should be aware of the subtle meanings of these feedback signals. As an example, human trainers tend to have a positive bias~\cite{Thomaz2007, thomaz08sociallyguided}. 

In addition, there are a variety of technical challenges to address when implementing RL in social robotics and social HRI.  One of the drawbacks of online learning through interaction with a human is the requirement of long interaction time, which can be tedious and impractical for the users, resulting in fatigue and a loss of interest. A considerable amount of interaction time can wear out the robot's hardware.  An alternative is using a simulated world to train the algorithm and subsequently deploying it on the real robot. Using a simulated setting has several advantages. It allows the agent to carry out learning repeatedly, which would otherwise be very expensive in the real-world. Simulated environments can also run much faster than the real-world, thus permitting the learning agent to make proportionately more learning experiences. Bridging the gap between the simulated and the real-world is not a simple task. It may be achieved by randomizing the simulator and learning a policy that shows success across many simulators and can ultimately be robust enough to work in the real world. However, simulating the real-world can be very difficult, especially with regards to modeling relevant human behaviors. Simulating the human requires a predictive model of human interactive behaviors and social norms as well as modeling the uncertainty of the real-world. Furthermore, the use of RL in social robotics poses other challenges such as devising proper reward functions and policies, as well as dealing with the sparseness of the reward signals. 

The exploration-exploitation dilemma is a well-known problem in RL and refers to the choice of actions to discover the environment or taking actions that have already proven to be effective in producing reward~\cite{rlbooknew}. RL practitioners use different approaches to deal with the trade-off between exploration and exploitation, such as epsilon-greedy policy~\cite{patompak2019learning}, epsilon-decreasing policy~\cite{chan2012} and Boltzmann distribution~\cite{perula2019bioinspired}. The~epsilon-greedy strategy exploits knowledge for maximizing rewards (greedily choosing the current best option), otherwise to select a random action with probability $\varepsilon \in [0,1]$~\cite{rlbooknew}. The epsilon-decreasing strategy decreases $ \varepsilon $ over time, thereby progressing towards exploitative behavior~\cite{rlbooknew}. Boltzmann exploration uses Boltzmann distribution to select the action to execute. A temperature parameter balances between exploration and exploitation (high-temperature value for selecting actions randomly and low-temperature value for selecting actions greedily)~\cite{rlbooknew}. 

Despite the mentioned challenges, there are also advantages of using RL in social robotics. One of the main advantages is that the robot can learn a personalized adaptation for different interactants, i.e., a different policy for each user. Social robots can learn social skills from their own actions without demonstrations through uncontrolled interaction experiences. This is especially true given that interaction dynamics are difficult to model and sometimes even humans cannot explain why they behave in a certain way. Therefore, RL may enable social robots to adapt their behaviors according to their human partners for natural human-robot interaction. In IRL, the immediate reward provided by the human teacher has the potential to improve the training by reducing the number of required interactions. Human teachers' guidance significantly reduces the number of states explored, and the impact of teacher guidance is proportional to the size of the state space, i.e., it increases as the size of the state space grows~\cite{suay2011effect}. In RL, how to achieve a goal is not specified, instead the goal is encoded and the agent can devise its own strategy for achieving that goal. Intrinsically motivated reward signals might be useful in many real-world scenarios, where sparse rewards make the goal-directed behavior challenging. Approaches using human social signals have the advantage of utilizing signals that the user exhibits naturally during the interaction. It does not require an extra effort to collect the reward. However, the change in social signals would not be so sudden, which would very much affect the time for convergence. The role of human social factors deserves extra attention in online learning methods. Combination of RL with deep neural networks has shown success in many application areas. DRL is also a trending technique in social robotics as we see increasing work in recent years. It has the advantage of not needing manual feature engineering~\cite{cuayahuitl2019data} and resulting in human-like behavior for social robots~\cite{Qureshi2016}.

\section{Reinforcement Learning}
\label{background}

\begin{figure}[!h]
\centering
\includegraphics[width=0.7\textwidth]{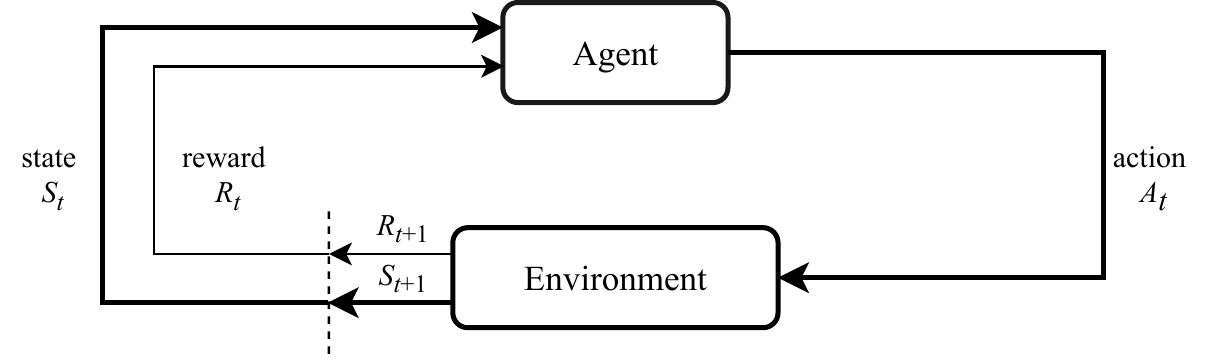}
\caption{Standard reinforcement learning framework (reproduced from~\citep[p.~38]{rlbooknew})}
\label{fig:rl_framework}
\end{figure}

Reinforcement learning~\citep{sutton1998introduction} is a framework for decision-making problems. Markov Decision Processes (MDPs) are mathematical models for describing the interaction between an agent and its environment. 
Formally, an MDP is denoted as a tuple of five elements $\langle$$\mathcal{S}$, $\mathcal{A}$, $\mathcal{P}$, $\mathcal{R}$, $\gamma$$\rangle$ where $\mathcal{S}$ represents the state space (i.e., the set of possible states), $\mathcal{A}$ represents the action space (i.e., the set of possible actions), $\mathcal{P}$ : $\mathcal{S}$$\times$$\mathcal{A}$$\times$$\mathcal{S}$ $\to$ $[0,1]$ represents the probability of transitioning from one state to another state given a particular action, $\mathcal{R}$ : $\mathcal{S}$$\times$$\mathcal{A}$$\times$$\mathcal{S}$ $\to$ $\mathbb{R}$ represents the reward function, and $\gamma$ is the discount factor that determines the importance of future rewards, $\gamma \in [0,1]$. The agent interacts with its environment in discrete time steps,~$t=0,1,2,..$; at each time step~$t$, the agent gets a representation of the environmental state~$S_t \in \mathcal{S}$, takes an action~$A_t \in \mathcal{A}$, moves to next state~$S_{t+1}$, and receives a scalar reward~$R_{t+1} \in \mathcal{R}$. Figure \ref{fig:rl_framework} depicts the standard RL framework. 

The agent's behavior that maps states to actions is described as a policy, $\pi : \mathcal{S} \times \mathcal{A}$ where $\pi(s|a) = Pr(A_t = a|S_t = s)$ is the probability of taking action $a \in \mathcal{A}$ given state $s$. 
The agent's goal is to maximize the expected cumulative discounted reward, in other words \emph{return} which is denoted as $G_{t}$:
\begin{equation}
    G_{t} = \sum_{k=0}^{\infty} \gamma^{k}R_{t+k+1}
\end{equation}
where $\gamma$ is the discount factor and usually $\gamma \in [0,1]$. The optimal behavior that is taking the best action at each state to maximize the reward over time is called \emph{optimal policy}, $\pi^*$.

\begin{figure}[!ht]
\centering
\includegraphics[width=0.7\textwidth]{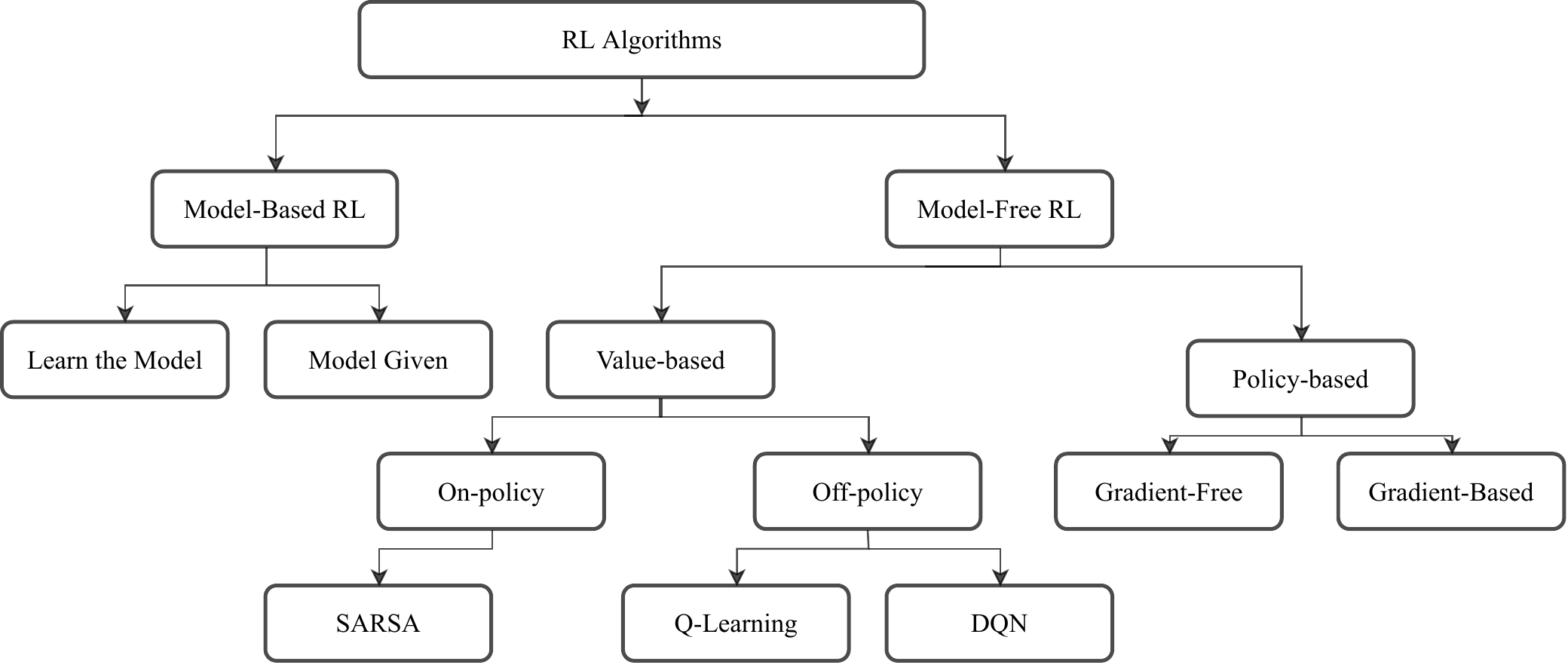}
\caption{Taxonomy of Reinforcement Learning algorithms (reproduced and shortened from~\citep{Zhang2020}).} 
\label{fig:rl_categories}
\end{figure}

There exists a large variety of approaches in RL. They can be most broadly distinguished as model-based and model-free. Model-free approaches can be further subdivided into value-based and policy-based approaches. Below, we briefly explain model-based and model-free RL in Section~\ref{model_based_model_free}, value-based methods in Section~\ref{value_based_methods} and policy-based methods in Section~\ref{policy_based_methods}.  A shortened version of a RL taxonomy can be seen in Figure~\ref{fig:rl_categories}.

\subsection{Model-Based and Model-Free Reinforcement Learning} 
\label{model_based_model_free}

RL algorithms can be divided into two main categories, model-free RL and model-based RL, depending on whether the agent does or does not use a model of the environment dynamics, which can be either provided or learned.
The model describes the transition function, $\mathcal{P}$, and the reward function, $\mathcal{R}$. The model-based methods can be divided into two categories: those that use a \textit{given model}, i.e., the models of the transition and the reward function can be accessed by the agent, and the methods in which the agent \textit{learns the model} of the environment~\citep{Zhang2020}. 
In the latter approach, the agent learns a model, which it subsequently uses during policy improvement. The agent can collect samples from the environment by taking actions. From those samples state transitions and reward can be predicted through supervised learning. Planning methods can be used directly on the environment model.

In the \qt{model-free} approach, there is no effort to build a model of the environment, instead the agent searches for the optimal policy through trial and error interactions with the environment. Model-free methods are easier to implement in comparison with model-based methods. These methods can be advantageous over more complex methods when building a sufficiently accurate model is difficult~\citep[p.~10]{rlbooknew}.

\subsection{Value-Based Methods}
\label{value_based_methods}

The value of policy $\pi$, namely the value function, is used to evaluate the states based on the total reward the agent receives over time. RL methods that approximate the value function through temporal difference (TD) learning instead of directly learning the policy $\pi$ are called \emph{value-based methods}. For each learned policy $\pi$, there are two related value functions: the state-value function, $v_\pi(s)$, and state-action value function (quality function), $q_\pi(s,a)$. The equations for $q_\pi(s,a)$ and $v_\pi(s)$ are given in Eq.~(\ref{eq2}) and Eq.~(\ref{eq1}) respectively.
$E_\pi$ in Eq.~(\ref{eq2}) and Eq.~(\ref{eq1}) means the agent follows policy $\pi$ in each step. 
\begin{equation}
\label{eq2}
    q_\pi(s,a) = E_\pi[R_{t+1}+ \gamma~R_{t+2} + \gamma^2~R_{t+3}+...|S_t = s, A_t = a] = E_\pi\left[\sum_{k=0}^{\infty} \gamma^{k}R_{t+k+1} \Bigg| S_{t} = s, A_{t} = a\right]
\end{equation}
\begin{equation}
\label{eq1}
    v_\pi(s) = E_\pi[R_{t+1}+ \gamma~R_{t+2} + \gamma^2~R_{t+3}+...|S_t = s] = E_\pi\left[\sum_{k=0}^{\infty} \gamma^{k}R_{t+k+1} \Bigg| S_{t} = s\right].
\end{equation}
The value functions are expressed via the Bellman equation~\citep{bellman1952theory}. The Bellman equation for $v_{\pi}$ and $q_{\pi}$ is given in Eq.~(\ref{eq3}) and Eq.~(\ref{eq4}) where $s^{\prime}$ indicates the next states from the set $\mathcal{S}$. 
\begin{equation}
\label{eq3}
    v_\pi(s) = \sum_{a} \pi(a|s) \sum_{s^{\prime}, r} p(s^{\prime}, r|s,a)[r + \gamma v_{\pi}(s^{\prime})]
\end{equation}
\begin{equation}
\label{eq4}
     q_\pi(s,a) = \sum_{s^{\prime}} p(s^{\prime}|s,a) \Bigg[ r(s, a, s^{\prime}) + \gamma\sum_{a^{\prime}} \pi(a^{\prime}|s^{\prime}) q_{\pi}(s^{\prime}, a^{\prime})\Bigg].
\end{equation}
Comparing policies, a policy $\pi$ is better than or equal to a policy $\pi^{\prime}$ if:
\begin{equation}
   \pi\geq\pi^{\prime} \textnormal{ if } \forall s \in \mathcal{S}:v_{\pi}(s)\geq v_{\pi^{\prime}}(s).
\end{equation}
There exists always at least one optimal policy $\pi^{*}$ whose expected return is greater than or equal to the other policy/policies for all states. Optimal policies share the same state-value function, defined as $v^{*}(s) = \underset{{\pi}}{max}~v_{\pi}(s)$ for all $s \in \mathcal{S}$, and action-value function, defined as $q^{*}(s,a) = \underset{\pi}{max}~q_{\pi}(s,a)$ for all $s \in \mathcal{S}$ and $a \in \mathcal{A}(s)$. The Bellman optimality equation for $q^{*}(s,a)$ is given in Eq.~(\ref{bellman_op}). 
\begin{equation}
\label{bellman_op}
    q^{*}(s,a) = \sum_{s^{\prime},r} p(s^{\prime},r|s,a) \Big[ r+ \gamma ~\underset{a^{\prime}}{max}~q^{*}(s^{\prime}, a^{\prime})\Big].
\end{equation}
Another distinction in RL methods comes from the perspective of policy: on-policy vs. off-policy learning. On-policy methods learn the value of the policy that is used to make decisions. In the on-policy setting, the target policy and the behavior policy are the same. The target policy is the policy that is learned about, and the behavior policy is the policy that is used to generate behavior. The state-action-reward-state-action (SARSA) algorithm~\citep{rummery1994line} is one of the on-policy methods in which the agent interacts with the environment, selects an action based on the current policy, then updates the current policy. The Q function update in SARSA is done using Eq.~(\ref{eq:sarsa_update}). A transition from one state-action pair to the next is expressed as $(S_{t}, A_{t}, R_{t+1}, S_{t+1}, A_{t+1})$ which gives rise to the name SARSA. The update given in Eq.~(\ref{eq:sarsa_update}) is done after every transition from a non-terminal state $S_{t}$.
\begin{equation}
\label{eq:sarsa_update}
    Q(S_{t}, A_{t}) \leftarrow Q(S_{t}, A_{t}) + \alpha \big[ R_{t+1} + \gamma~Q(S_{t+1}, A_{t+1}) - Q(S_{t}, A_{t}) \big].
\end{equation}
In the off-policy methods, the target policy is different from the behavior policy. In these methods, the policy that is evaluated and improved does not match the policy that is used to generate data. Off-policy methods can re-use the experience from old policies or other agents' interaction experience to improve the policy. One example of an off-policy algorithm is Q-learning~\citep{watkins1989learning}. It is one of the most popular RL algorithms using discounted reward~\citep{gosavi2006boundedness}.
The Q-learning rule is defined by:
\begin{equation}
\label{q_learning}
    Q(S_{t}, A_{t}) \leftarrow Q(S_{t}, A_{t}) + \alpha \big[ R_{t+1} + \gamma~\underset{a}{max}~Q(S_{t+1}, a) - Q(S_{t}, A_{t}) \big].
\end{equation}
The Q-learning algorithm iteratively applies the Bellman optimality equation (given in Eq.~(\ref{bellman_op})). 
As shown in Eq.~(\ref{q_learning}), the
main difference between Q-learning and SARSA (see Eq.~(\ref{eq:sarsa_update})) is that in the former the target value is not dependent on the policy being used and only depends on the state-action function. The Q-learning algorithm is given in Algorithm~\ref{q_learning_algo} where an episode (or trial) describes a sequence from the initial state to the terminal state, and $\alpha$ specifies a learning rate (controls the amount of new information).
Q-learning, along with its different variations, is the most commonly used RL method in social robotics.

\subsection{Policy-Based Methods}
\label{policy_based_methods}
Policy-based methods, also known as direct policy search methods, do not use value function models. In these methods, the policy is parameterized with $\theta$ and written as $\pi_{\theta}$. They operate in the space of policy parameters $\Theta$ and $\theta \in \Theta$~\citep{deisenroth2013survey}. The goal is still to maximize the accumulative return. The agent updates its policy by exploring various behaviors and exploiting the ones that perform well in regard to some predefined utility function $J(\theta)$. In many robot control tasks the state space, which includes both internal states and external states, is high-dimensional. The policy of the robot $\pi_{\theta}$ can be defined as a controller. For any state of the robot, this controller decides which actions to take or which signals to send to the actuators~\citep{sigaud2019policy}. The robot takes its actions $u$ according to the controller (please note, actions in policy search context are represented with $u$ instead of $a$). The robot controller can be stochastic, i.e., $\pi(u|s)$ or deterministic, i.e., $\pi(s)$. After the action execution the robot transitions to another state according to the probabilistic transition function $p(s_{t+1}|s_t, u_t)$. These states and actions of the robot form a \emph{trajectory} $\tau = (s_0, u_0, s_1, u_1,...)$. The corresponding return for the trajectory $\tau$ is represented as $R(\tau)$. The global utility of the robot is denoted as:  
\begin{equation}
    J(\theta) = \mathbb{E}_{\tau \sim \pi_{\theta}}[R(\tau)].
\end{equation}
Computing the expectation in $\mathbb{E}_{\tau \sim \pi_{\theta}}[R(\tau)]$ requires to run an infinite number of trajectories with the current controller. The way to go around this difficulty is to sample the expectation. After performing a finite set of trajectories, the return is computed over these trajectories. Thus, the goal is:
\begin{equation}
    \theta^{*} = \underset{\theta}{\argmax}~J(\theta) = \underset{\theta}{\argmax} \sum_{\tau} P(\tau, \theta) R(\tau)
\end{equation}
where $\theta^{*}$ is the estimate of global performance and $P(\tau, \theta)$ is the probability of $\tau$ under policy $\pi_{\theta}$.  

Here RL addresses a black-box optimization problem in that the function which relates the performance to the policy parameters is unknown. There are two families of methods: direct policy search and gradient descent~\citep{sigaud2019policy}. In direct policy search algorithms, approximate gradient descent is performed by \qt{random trial then selection} methods, like genetic algorithms, evolution strategies, finite differences, cross entropy, etc. These algorithms need many samples and can escape from local minima if large enough variations are used. In gradient descent methods, a mathematical transformation is used so that policy gradient methods can be applied. In these methods, the policy gradient update is given by:
\begin{equation}
\label{gradient_update}
    \theta_{k+1} = \theta_{k}+\alpha \nabla_{\theta} J(\theta)
\end{equation}
where $\alpha$ is a learning rate, and the policy gradient is given by~\citep{deisenroth2013survey}:
\begin{equation}
   \nabla_{\theta} J(\theta) = \sum_{\tau} \nabla_{\theta} P(\tau, \theta) R(\tau).
\end{equation}
There are different methods to estimate the gradient $\nabla_{\theta} J(\theta)$, interested readers may refer to~\citep{deisenroth2013survey}.
Policy-based methods have the advantage of being effective in high dimensional or continuous action spaces and having better convergence properties.

Some methods learn both policy and value functions. These methods are called \emph{actor-critic methods}, where \qo{actor}
is the learned policy that is trained using policy gradient with estimations from the critic, and \qo{critic} refers to the learned value function that evaluates the policy. 

\subsection{Deep Reinforcement Learning}
Learning in RL progresses over discrete time steps by the agent interacting with the environment. Obtaining an optimal policy requires a considerable amount of interaction with the environment, which results in high memory and computational complexity. Therefore, the tabular approaches that represent state-value functions, $v_\pi(s)$, or state-action value functions, $q_\pi(s,a)$, as explicit tables are limited to low-dimensional problems, and they become unsuitable for large state spaces. A common way to overcome this limitation is to find a generalization for estimating state values by using a set of features in each state. In other words, the idea is to use a parameterized functional form with weight vector $w\in\mathbb{R}^d$ for representing  $v_{\pi}(s)$ or $q_{\pi}(s,a)$ that are written as $\hat{v}(s; \theta)$ or $\hat{q}(s,a;\theta)$ instead of tables~\citep[p.~161]{rlbooknew}. Such approximate solution methods are called \emph{function approximators}. The reduction of the state space by using the generalization capabilities of neural networks, especially deep neural networks, is becoming increasingly popular. Deep Learning (DL) has the ability to perform automatic feature extraction from raw data. Deep Reinforcement Learning (DRL) introduces DL to approximate the optimal policy and/or optimal value functions~\citep[p.~192]{rlbooknew}. Recently, there has been an increasing interest in using DL for scaling RL problems with high-dimensional state spaces.

The DQN method, first presented by~\citeauthor{mnih2015human}, combines Q-learning with convolutional neural networks for learning to play a wide variety of Atari games better than humans~\citep{mnih2015human}. 
In DQN, the agent's experiences $e_t = (s_t, a_t, r_t, s_{t+1})$ are stored at each time step $t$ in a data set $D_t = \{e_1,...,e_t\}$, so-called \emph{experience replay memory}. Q-learning updates are applied on a mini-batch uniformly sampled from the experience replay memory.
The Q-learning update is done using Eq.(\ref{dqn_lossfn}):
\begin{equation}
    \label{dqn_lossfn}
     L_{i}(\theta_{i}) = \mathbb{E}_{s,a,r,s^\prime\sim U(D)}\Big[ \Big( r + \gamma~\underset{a^\prime}{max}~Q(s^\prime, a^\prime; \hat{\theta}_i) - Q(s,a; \theta_i)\Big)^2 \Big]
\end{equation}
where $\theta_i$ represents the parameters (weights) of the Q-network at iteration $i$ and $\hat{\theta}_i$ represents the parameters used to compute the target network at iteration $i$. The target network parameters $\hat{\theta}_i$ are updated to the parameters $\theta_i$ after every $C$ iterations. 
\section{Categorization of RL Approaches in Social Robotics Based on RL Type}
\label{categories_rl_based}

In human-human communication, a communication medium is a means of conveying information to other people. It can be in different forms such as verbal, nonverbal, affective, and tactile. Human-robot interaction overlaps with human-human interaction to a certain extent. Furthermore, there can be an additional physical interface (i.e., a computer, a tablet, a smart game board, etc.) shared between the robot and the human. In the interaction between the robot and the human, information transmission is bidirectional, the robot and the human can be sender, receiver, or both. In the surveyed papers, we see all these communication channels being utilized, especially for the RL problem formulation. As it has already been stated in the introduction, one of the prominent characteristics of social robots is the ability to interact and communicate. Therefore, we provide two categorizations in this section: first we categorize the papers based on RL types, after which we provide a further discussion and categorization with respect to the utilized communication channels and interaction dynamics for the reward functions. 

\subsection{Bandit-Based Methods}

Bandit-based methods can be considered as a simplified case of RL in which the next state does not depend on the action taken by the agent. Different bandit-based methods explored in social robotics~\cite{schneider2017exploring, ritschel2019adaptive, leite2011modelling, ritschel2018drink, gao2018robot}, such as dueling bandit learning~\citep{schneider2017exploring}, k-armed bandit method (multi-armed bandit)~\cite{ritschel2019adaptive, leite2011modelling, ritschel2018drink}, and Exponential-Weight Algorithm for Exploration and Exploitation (Exp3) algorithm~\citep{gao2018robot}. 

\subsubsection{{Additional Physical Communication Medium between the Robot and the Human}} 
Learning user preferences to personalize the user experience is used in customizing advertisements and search results. A similar approach was applied in HRI studies~\cite{schneider2017exploring, ritschel2019adaptive}. Whereas the customization is done in the background for personalized experiences in websites using users' clicks, it is adapted for social interactions by asking the user to select their preferences using the buttons. In other words, these studies use a physical communication medium between the robot and the human.~\citet{schneider2017exploring} investigated a dueling bandit learning approach for preference learning. The algorithm draws two or more actions, and the relative preference is used as reward. It is defined as follows: In each time step $t>0$ a pair of arms $(k^{(1)}_{t}, k^{(2)}_{t})$ is selected and presented to the user, if the user prefers $k^{(1)}_{t}$ over $k^{(2)}_{t}$ then $w_t = 1$, and $w_t = 2$ otherwise where $w_t$ is a noisy comparison result. The distribution of outcomes is represented by a preference matrix $P = [p_{ij}]_{KxK}$, here $p_{ij}$ is the probability that the user preferred arm $i$ over arm $j$. The~participant provided pairwise comparisons via a button. In the work by~\citet{ritschel2019adaptive}, the robot adapted its linguistic style to the user's preferences. They~defined the learning tasks as k-armed bandit problems. The adaptation was done based on explicit human feedback given via buttons in the form of numeric reward ($-$1, +1). The actions of the robot were a set of scripted utterances. Similarly,~\mbox{\citet{ritschel2018drink}} used an additional medium between the robot and the user. They employed the social robot Reeti as a nutrition adviser, where~a custom hardware was utilized to obtain the information about the selected drink~\cite{ritschel2018drink}. Their custom hardware included an electronic vessel holder and a smart scale that could communicate with the robot. The problem was formalized as an k-armed bandit problem where the actions of the robot were scripted spoken advice. The reward was calculated from the amount of calories and quantity of the selected drink. 

\subsubsection{{Verbal and Nonverbal Communication Plus an Interface}}
Social robots can use any natural communication channel, and benefit from different user interfaces. The studies~\cite{leite2011modelling, gao2018robot, ritschel2018drink} take advantage of a physical medium shared across the robot and the human to simplify the state space representations.~\citet{leite2011modelling} used a multi-armed bandit for empathetic supportive strategies in the context of a chess companion robot for children. The difference in the probabilities of the user being in a positive mood before and after employing supportive strategies was used as a reward. The child's affective state was calculated by using visual facial features (smile and gaze) and contextual features of the game (game evolution i.e winning/losing, chessboard configuration). Similarly, in the work by~\citet{gao2018robot} the user's task-related parameters were monitored through the puzzle interface. The robot's behaviors were adapted by combining a decision tree model with the Exp3~\citep{bubeck2012regret}. The Exp3 algorithm maintains a list of weights for each of the actions, which are used for selecting the next action. The reward was the user's task performance in combination with the user's verbal feedback. The set of robot actions included four supportive behaviors to help the user to solve the puzzle game.

\subsection{Model-Based and Model-Free Reinforcement Learning} 
\label{model_based_model_free}

\subsubsection*{{Verbal Communication}}
	Considering the challenge of modeling real-world human-robot interactions, the~majority of papers included in this survey use model-free RL. Nevertheless, several recent works started to investigate model-based RL for HRI~\cite{Tseng18, martins2019}. One of the challenges of real-world robot learning is the delayed reward. There is an assumption that the result of an agent's observations of its environment is available instantly. However, there can be a lag in human reaction to robot actions in HRI. When the reward of the robot depends on human responses, reward shaping can be useful for the robot to get more frequent feedback. Reward shaping is a technique that consists of augmenting the natural reward signal so that additional rewards are provided to make the learning process easier~\cite{Wiewiora2010}. Studies in~\cite{Tseng18, martins2019} presented methods including model-based RL and reward shaping for HRI.~\citet{Tseng18} proposed a model-based RL strategy for a service robot learning the varying user needs and preferences, and adjusting its behaviors. The proposed reward model was used to shape the reward through human feedback by calculating temporal correlations of robot actions and human feedback. Concretely, they modeled human response time using a gamma distribution. This formulation was found to be effective (more cumulative reward collected) in dealing with delayed human feedback. The~work by~\citet{martins2019} presented a user-adaptive decision-making technique based on a simplified version of model-based RL and POMDP formulation. Three different reward functions were formulated, and compared in the experiments. Their entropy-based reward shaping mechanism devised using an information-based term. The purpose of using the information term was to increase the reward given for an action leading to unknown transitions, thereby encouraging the robot to investigate the impact of new actions on the~user. 

\subsection{Value-Based Methods}

In recent years, there has been an increasing interest in applying RL methods to social robotics with growing trend towards value-based methods. Q-learning, along with its different variations, is the most commonly used RL method in social robotics. The studies using Q-learning are~\cite{Barraquand2008, yang2017, zarinbal2019new, addo2014applying, Chiang2015, Ritschel2017b, park2019model, Weber2018a, maroto2018bio, perula2019bioinspired, papaioannou2017hybrid, hemminghaus2017, moro18}. These comprise studies using standard Q-learning~\cite{zarinbal2019new, addo2014applying, park2019model, nejat2008can, papaioannou2017hybrid, hemminghaus2017}, studies modify Q-learning for dealing with delayed reward~\cite{Barraquand2008}, tuning the parameters for Q-learning such as $\alpha$~\cite{Barraquand2008, perula2019bioinspired, maroto2018bio}, dealing with decreasing human feedback over time~\cite{Barraquand2008}, comparing their proposed algorithm with Q-learning~\citep{Tseng18, Silva12, moro18, chan2012, castro2014learning}, variation of Q-learning called Object Q-learning~\citep{castro2014learning, castro2013autonomous, Castro-Gonzalez2011}, combining Q-learning with fuzzy inference~\cite{Chen2018}, SARSA~\cite{gordon2016affective, Gamborino18}, TD($\lambda$)~\cite{Ranatunga2011}, MAXQ~~\citep{chan2012, chan2011minimizing, chan2011learning}, R-learning~\cite{patompak2019learning}, and
Deep Q-learning~\citep{Qureshi2016, qureshi2017show, Qureshi2018, cuayahuitl2019data}. 

\subsubsection{{Tactile Communication}}
When the user is involved in the learning process by providing feedback in the form of reward or guidance, the general approach is either using an additional interface or utilizing the sensory information such as internal (robot's onboard sensors) or external cameras and microphones. Nowadays, many social robots are equipped with tactile capabilities. However, the usage of the robots' touch sensors as a feedback mechanism has received relatively little attention in the context of RL in social robotics. Yet~\cite{Barraquand2008,yang2017} benefited from the robot's tactile sensors instead of an additional interface between the user and the robot.~\citet{Barraquand2008} conducted five experiments with different modifications of the classical Q-learning algorithm. The human teacher provided feedback through tactile sensors of the Sony AIBO robot, caressing the robot for the positive feedback and tapping the robot for the negative feedback. The action space comprised two actions; bark and play. The first experiment was standard Q-learning with human reward. Since the human ceased giving feedback over time, they concluded that the learning rate $\alpha$ should be adapted. In the second experiment, they used the asynchronous Q-learning algorithm. In asynchronous Q-learning, the learning rate $\alpha$ may be different for different state-action pairs. The learning rate is decreased when the system encountered the same situations and actions. In relation to standard Q-learning this modification increased the effectiveness of the algorithm, i.e., it~learned faster and forgot more slowly. Because the learning rate was much smaller when there was no feedback. To overcome the delayed reward, they considered to increase the effect of human-delivered positive reward in larger time frames and to decrease the effect of negative reward in a shorter time frame. The use of an eligibility trace with a heuristic for delayed reward was found to be more efficient than classical Q-learning (generalizing experience to cover similar situations). The authors noted that learning rate, reward propagation, and analogy (i.e., propagating information to similar states) can improve the effectiveness of learning from social interaction.~\mbox{\citet{yang2017}} proposed a Q-learning based approach that combines homeostasis and IRL. The internal factors, i.e., the~drives and motivations worked as a triggering mechanism to initiate the robot's services. However, the reward in the real-world experiments was given by the user touching the robot's head, left hand, and right hand to give positive, negative, and dispensable feedback, respectively~\citep{yang2017}. The authors trained their model in a simulator and deployed it on the Pepper robot. 

\subsubsection{{Additional Physical Communication Medium between the Robot and the Human}}
Since we identify social robots with interaction, the robot learning within a social scenario stands out in the surveyed papers. Alternatively, there are studies where social interaction is not the main concern however, the main purpose is training a social robot to do a task. As an example, a human teacher trains the agent through a GUI~\cite{suay2011effect, suay12}, speech~and gestures~\cite{Thomaz2007, thomaz08sociallyguided}.
In ~\citet{suay2011effect}, human teacher trained a social robot. They~performed experiments similar to those presented in~\cite{thomaz2006adding} in a real-world scenario with the Nao robot~\cite{suay2011effect}. The human trainer observed the robot in its environment via a webcam and provided reward based on the robot's past actions or anticipatory guidance for selecting future actions through a GUI. They conducted four sets of experiments (small state space and only reward, large state space and only reward, small state space and reward plus guidance, large state space and reward plus guidance) to investigate the effect of teacher guidance and state space size on learning performance in IRL. The task was object sorting and the size of state space depended on the object descriptor features. Their~results showed that the guidance accelerated the learning by significantly decreasing the learning time and the number of states explored. They observed that human guidance helped the robot to reduce the action space and its positive effect was more visible in large state-space. In a similar vein,~\citet{suay12} conducted a user study in which 31 participants taught a Nao robot to catch the robotics toys by using one of three algorithms: Behavior Networks, IRL, and Confidence-Based Autonomy. The study compared the results of these algorithms in terms of algorithm usability and teaching performance by non-expert users. In IRL, the participants provided positive or negative feedback in the form of reward through an on-screen interface. In terms of teaching performance, users achieved better performance using Confidence-Based Autonomy, however, IRL was better of modelling user behavior. It has been noted in much of the literature that teaching with IRL requires more time than with other methods because users had the tendency to stop rewarding or to vary their reward strategy. This affected the training time, which is a drawback to this~approach. 

\subsubsection{{Verbal and Nonverbal Communication}}
We discuss different human feedback types in IRL in Section~\ref{interactive_rl}. When a human teacher trains an agent, the positive or negative feedback might convey several meanings, even lack of feedback can give information to the agent depending on the teacher's training strategy~\cite{loftin2016learning}. For example,~\citet{Thomaz2007} realized that human trainers might have multiple intentions with the negative reward they are giving, such as the last taken action was bad and future actions should correct the current state. They performed experiments with two different platforms: the Leonardo robot learned pressing buttons and a virtual agent learned baking a cake (Sophie's kitchen). The virtual agent responded to the negative reward by taking an UNDO action, i.e., the opposite action. In the examples with the Leonardo robot, the human teacher provided verbal feedback. After negative feedback, the robot expected the human teacher to guide it through refining the example by using speech and gestures (collaborative dialog). Although the interactive Q-learning with the addition of UNDO behavior was tested only on the virtual agent, it is worth mentioning that the proposed algorithm was more efficient compared to standard IRL. It had several advantages such as robust exploration strategy, fewer states visited, fewer~failures occurred and fewer action trials done for learning the task. Continuing along these lines, ~\citet{thomaz08sociallyguided} explored how self-exploration and human social guidance can be coupled for leveraging intrinsically motivated active learning. They called the presented approach socially guided exploration, in which the robot could learn by intrinsic motivations, however, it could also take advantage of a human teacher's guidance when available. The robot learner with human guidance generalized better to new starting states and reached the desired goal states faster than the self-exploration. 

\subsubsection{{Higher Level Interaction Dynamics: Engagement}}
	Social robots are expected to exhibit flexible and fluent face-to-face social conversation. The natural conversational abilities of social robots should not be only limited to short basic task related sentences. However, they should be able to engage users in the interactions with chat and entertainment, varying from storytelling to jokes together with human-like vocalizations and sounds. As an example,~\citet{papaioannou2017hybrid} reported that users spent more time with the robot which can carry out small chat together with task-based dialogue compared to the robot that conversed only task-based dialogue. In their system, the agent was trained using the standard Q-learning algorithm with simulated users and tested with the Pepper robot where the robot assisted visitors of a shopping mall by providing information about and directions to the shops, current discounts in the shops, among other things. In the problem definition, states were represented with 12 features such as user engaged, task completed, distance, turn taking, etc. The action space consisted of 8 actions, $A = $ [PerformTask, Greet, Goodbye, Chat, GiveDirections, Wait, RequestTask, RequestShop]. The~reward was encoded as predefined numerical values based on task completion by the agent, including the engagement of the user. Another study considering user engagement is~\citet{Keizer2014}, who applied a range of ML techniques in the presented system that included a modified iCat robot (with additional manipulator arms with grippers) and multimodal input sensors for tracking facial expressions, gaze behavior, body language and location of the users in the environment. The reward function was a weighted sum of task-related parameters. For each individual user $i$ the reward function $R_i$ was defined as $R_i = 350 \times TC_i - 2 \times W_i - TO_i - SP_i$.~~$TC_i$ is short for Task Complete, and is a binary variable. $W_i$ (Waiting) is a binary variable showing whether the user $i$ is ready to order but not engaged with the system.~~$TO_i$ stands for Task Ongoing and is a binary variable describing whether the user is interacting with the robot but has not been served. $SP_i$~is short for Social Penalties and corresponds to several social penalties (e.g., while the user $i$ is still talking to the system, it turns its attention to another user). An experimental evaluation compared a hand-coded and trained system. The authors reported that the trained system performed better and it was found to be faster at detecting user engagement than the hand-coded one, while the latter was more stable. In~\cite{Ritschel2017b, Weber2018a, addo2014applying}, the authors investigated the entertainment capabilities of social robots using RL.~\citet{Ritschel2017b} presented a social-cues-driven Q-learning approach for adapting the Reeti robot to keep the user engaged during the interaction. The engagement of the user was estimated from the user's movement through the Kinect 2 sensor by using a Dynamic Bayesian Network. They used the change in the engagement as a reward in the storytelling scenario to adapt the robot's utterance based on the personality of the user. In similar fashion, the work by~\citet{Weber2018a} incorporated social signals in the learning process, namely the participants' vocal laughs and visual smiles as reward. In the problem formulation, they used a two-dimensional vector containing probabilities of laughs and smiles for state representation, and the action space consisted of sounds, grimaces and three types of jokes. They used an average reward based on all samples from the punchline to the end with a predefined punchline for every joke. The human social signals were captured and processed by using the Social Signals Interpretation (SSI) framework~\cite{wagner2013social}. Their purpose was to understand the user's humor preferences in an unobtrusive manner in order to improve the engagement skills of the robot. In a joke-telling scenario, the Reeti robot adapted its sense of humor (grimaces, sounds, three kinds of jokes and their combination) by using Q-learning with a linear function approximator. Likewise~\citet{addo2014applying} presented a joke telling scenario with a torso Nao robot for entertaining a human audience. They used Q-learning in which the actions of the robot were pre-classified jokes, and the numerical reward corresponded to affective states of the user. However, the~affective states of the participants were captured by a self-reported feedback signal. After~each joke, the human participant provided a verbal feedback (i.e., reward) such as~\qt{very funny},~\qt{funny},~\qt{indifferent} and~\qt{not funny}.  

\subsubsection{{Affective Communication: Facial Expressions}}
Human facial expressions are perhaps one of the richest and most powerful tools in social communication. Facial expressions analysis is commonly used in HRI for understanding users and enhancing their experience. Affective facial expressions can also facilitate robot learning in RL. Recently, it is becoming more popular to use off-the-shelf applications in social robotics for different perception and recognition modules. Affectiva software~\cite{mcduff2016affdex} analyzes facial expressions from videos or in real-time. The studies~\cite{gordon2016affective, park2019model, Gamborino18} used this software for affective child-robot interaction. In the work by ~\citet{gordon2016affective} a tutoring system for children was presented. The system included an Android tablet and the Tega robot setup integrated with the Affectiva software for facial emotion recognition. They used the SARSA algorithm where the reward was a weighted sum of valence and engagement. Both valence and engagement values were obtained from the Affectiva software. Similar to ~\cite{gordon2016affective},~\citet{park2019model} used the Tega robot as a language learning companion for young children. A personalized policy was trained through 6--8 sessions of interaction by using a tabular Q-learning algorithm. The reward function was a weighted sum of engagement and learning gains of the child. The engagement was obtained from the Affectiva software. The learning gains in the reward function was represented as numerical values ([$-$100, 0, +50, +100]) depending on the lexical and syntactic complexity of the phrase relative to the child's level.~\citet{Gamborino18} presented an approach for socially assistive robots for children to support them in emotionally difficult situations using SARSA. In the proposed method, the human trainer selects the actions for the social robot RoBoHoN (small humanoid smartphone robot) through an interface with the purpose to improve the mood of the child depending on her/his current affective state. The affective state of the child was based on seven basic facial emotions and engagement obtained by the Affectiva software and stored in an input feature vector to classify the mood of the child as good or bad. The emotions were binarized as 1 or 0 depending on whether the value was greater or less than the average, respectively. The robot suggested a set of actions to the trainer. The aim was to suggest actions that would match with the trainer's action preferences. This way the agent would act independently, without feedback from the trainer.
Another study using facial expressions is~\citet{zarinbal2019new}, in which Q-learning was used for query-based scientific document summarization with a social robot. The problem formulation was as follows: In each state~$S_t:<x_i, score^t(x_i)>$~a summary that consisted of M sentences was generated, where $x_i$ is a sentence and $i = 1,2,..., M$. The scoring scheme was updated based on the human-delivered reward. The reward $r_t \in \{-1, 0, 1\}$ depended on the classified facial expressions: dislike, neutral and like. In~state $S_t$, the robot presented the sentence $x^*$ to the user and based on his reward $r_t$. The~authors concluded that user feedback may improve the query-based text summarization.

\subsubsection{{Verbal Communication}}
The curse of dimensionality is a phenomenon that refers to problems with high dimensional data. Representing state and action spaces as explicit tables becomes impractical for large spaces. To overcome the problem of large state space, approximate solutions are used, one of them being fuzzy techniques. This approach is also explored for HRI, e.g.,~\mbox{\citet{Chen2018}} and \citet{patompak2019learning} used fuzzification and fuzzy inference together with Q-learning. These works employed verbal communication in their user studies.~\citet{Chen2018} proposed a multi-robot system for providing services in a drinking-at-a-bar scenario. The authors used a modified Q-learning algorithm combined with fuzzy inference which was called information-driven fuzzy friend-Q (IDFFQ) learning for understanding and adapting the behaviors of the mentioned multi-robot system based on the emotion and intention of the user. The reward function was defined as $r = (r_t + r_h)/2$. Task completion $r_t$ (i.e., robots selected the drink the user preferred) and the human's satisfaction with the robots' task performance $r_h$ were predefined numerical values. Fuzzification of emotions was done using the triangular and trapezoidal membership function in the pleasure-arousal plane. They compared the proposed algorithm with their previous algorithm, Fuzzy Production Rule-based Friend-Q learning (FPRFQ)~\citep{chen2013adapting}. The authors noted that the current algorithm was superior in that it resulted in higher collected reward and faster response time of the robots.~\citet{patompak2019learning} proposed a dynamic social force model for social HRI. The authors considered two interaction areas: a quality interaction area and a private area. The quality interaction area was defined as the distance from which the users can be engaged in high-quality interactions with robots. The proposed model was designed by a fuzzy inference system, the membership parameters were optimized by using the R-learning algorithm~\cite{schwartz1993reinforcement}. R-learning is an average reward RL approach; it does not discount future rewards~\cite{mahadevan1996average}. They argued that R-learning was suitable for the scenario since they intended to take every interaction experience into account equally. In the real robot experiments, positive or negative verbal rewards were provided by the participants.

Another study that used verbal communication for the reward is~\cite{nejat2008can}. In this study, a gesture recognition system categorized the body trunk patterns as towards (the person is facing the robot), neutral (the trunk is facing the robot between $3^{\circ}$--$15^{\circ}$ away), and away ( orientation of the trunk is more than $15^{\circ}$). The recognized gestures were interpreted as a person's accessibility level, which was used to determine the person's affective state. In the Q-learning-based decision-making system, the robot had drives and emotional states which were utilized for action selection. In particular, a state is represented as $s(y_{H}, y_{R}, d)$ where $y_{H}$ is the accessibility level of the human, $y_{R}$ is emotional state of the robot and $d$ is the dominant drive. State transition probabilities, Q-values for each state, and reward for each transition were predetermined numerical values. The satisfaction of the robot's drives depended on the robot completing the task. In the experimental scenario, the Brian robot reminded the user about daily activities (eat, use the bathroom, go for a walk and take medication) and the user verbally stated \qo{yes} or \qo{no} after the robot's action, with \qo{no} meaning that the robot's drive is not satisfied and it will continue to try to satisfy the drive. The authors mentioned that the robot could use its drives in one or two iterations for the reminders except the drive related to using the bathroom. It was attributed to people potentially being uncomfortable with this reminder.

\subsubsection{{Higher Level Interaction Dynamics: Attention}}
Social robots have the potentials for information acquisition from both verbal and nonverbal communication. Not only can they gesture, maintain eye contact, and share attention with their users, but they can also estimate the users' non-verbal cues and behave accordingly. In this interaction, both actors can interpret verbal and nonverbal social cues to communicate effectively. For natural fluid HRI, robot non-verbal behaviors together with verbal communication are thoroughly discussed in~\cite{ mavridis2015review}. These social cues do not only convey a basic message but also carry higher-level interaction dynamics such as attention, engagement, comfort, and so on. The following works highlight these in the context of RL in social robotics.~\citet{Chiang2015} proposed a Q-learning based approach for personalizing the human-like robot ARIO's interruption strategies based on the user's attention and the robot's belief in the person's awareness of itself. The authors called it the \qt{robot's theory of awareness}. They formulated the problem based on the user attention, which was referred to as a Human-Aware Markov Decision Process. The human attention was estimated with a trained Hidden Markov Model (HMM) from human social cues (face direction, body direction, and voice detection). The reward consisted in predefined numerical values based on the robot's theory of awareness of the user. The robot had six actions (gestures: head shake and arm wave; navigation: approach the user and move around; audio: make sound and call name) to draw the user's attention while the user was reading. The optimal policy converged after two hours of interaction. The robot developed personalized policies for each user depending on their interruption preferences. Another study considering human attention in their problem formulation is~\citet{hemminghaus2017}. They used Q-learning to adapt the robot head Furhat's behavior in a memory game scenario. In the game, the robot assisted the participant by guiding their attention towards target objects in a shared spatial environment. In the proposed hierarchical approach, the high-level behavior was mapped to low-level behaviors, which could then be directly executed by the robot. The purpose of using Q-learning was to learn the execution of high-level behaviors through low-level behaviors. In the problem formulation, states were represented in terms of the user's gaze, user's speech, and game state. The game state represented the number of remaining card pairs in the game. The action space included actions such as speaking, gazing, etc. or a combination of those actions. The reward was designed as $r = r_{pos} - c ~~\textnormal{if success}~~r = c.r_{neg}~~\textnormal{if no effect}$. The robot received a positive reward $r_{pos}$ if the robot's action helped the user to find the correct pair. The robot received a negative reward $r_{neg}$ if the action had no effect on helping the user. $c$ represents the cost of the chosen action in cases where the costs were determined manually.~\citet{moro18} is another study that considered the attention of the user. Their scenario was an assistive tea-making activity for older people with dementia. The authors proposed an algorithm involving Learning from Demonstration (LfD) and Q-learning for personalized robot behavior according to a user's cognitive abilities~\citep{moro18}. The Casper robot learned to imitate the combination of speech and gestures from a collected data set. The robot learns to select the suitable labeled behavior (i.e., speech and gestures initially learned from demonstrations) that is most likely to transition the user into the desired state, i.e., focused on the activity and completing the correct step. The reward function, $R(s, b_l^{i})$, depended on $b_l^{i}$, the labeled behavior displayed by the robot, and the state $s$ where $s = \{ s_r, s_u\}$. Here, $s_r$ represents a set of robot activity states, and $s_u$ is the user state such that $s_u = \{ s_{fnc}, s_{ac}\}$. In the user state, $s_{fnc}$ represents the user functioning state which is one of five mental functioning states: focused, distracted, having a memory lapse, showing misjudgment, or being apathetic. The user activity state, $s_{ac}$, represents possible actions that can be performed by seniors with cognitive impairment: successfully completing a step, being idle, repeating a step, performing a step incorrectly, or declining to continue the activity. The robot was rewarded according to the state the user transitioned into---a positive reward if the user was focused and completed the activity, and a negative reward if the user transitioned to an undesirable state. The authors compared the proposed approach with Q-learning, and reported that the proposed approach required fewer interactions for convergence and fewer steps required to complete the tea-making activity. In all the papers explained above, the robot takes the users attention into account for deciding its actions. Shared attention refers to situations involving mutual gaze, gaze following, imperative pointing and declarative pointing.~\citet{Silva12} presented a robotic architecture for shared attention which included an artificial motivational system driving the robot's behaviors to satisfy its intrinsic needs, so-called necessities. The motivational system comprised necessity units that were implemented as a simple perceptron with recurrent connections. The input to the artificial motivational system was provided by a perception module used to detect the environmental state and to encode the state in first order logic with predicates. This module included face recognition with head pose estimation and a visual attention mechanism. The necessities of the robot were associated with a state-action pair in the training phase of the learning algorithm. The activation of a necessity unit was dependent on the input signal representing a stimulus detected from the environment (i.e., the perception module) and empirically defined parameters. They~compared the performance of three different RL algorithms, namely contingency learning, Q-learning and Economic TG (ETG) methods for shared attention in social robotics. ETG is a relational RL algorithm that incorporates a tree-based method for storing examples~\citep{da2009relational}. Because ETG performed better in the simulation experiments, they decided to employ it in real-world experiments which entailed one of the authors interacting with the robotic head. The authors reported that the robot's corrected gaze index, which was defined as frequency of gaze shifts from the human to the location that the human is looking at, was increased over time during learning.

\subsubsection{{Affective Communication}}
Humans use affective communication consciously or unconsciously in their daily conversations by expressing feelings, opinions, or judgments. Social robots can facilitate their learning process through sensing and building representations of affective responses. This idea was used in~\citep{chan2012, chan2011minimizing, chan2011learning}. 
In these studies, the socially assistive robot Brian 2.0 was employed as a social motivator by giving assistance, encouragement, and celebration in a memory game scenario. In the scenario, the participants interacted with the robot one-on-one with the objective to find the matching pictures in the memory card game (4~$\times$ 4 grid, 16 picture cards). The robot's behaviors were adapted using a MAXQ method to reduce the activity-induced stress in the user. The MAXQ approach is a hierarchical formulation, which accommodates a hierarchical decomposition of the target problem into smaller subproblems by decomposing the value function of an MDP into combinations of value functions of smaller integral MDPs~\citep{dietterich2000hierarchical}. The authors argued that the MAXQ algorithm was suitable for memory game scenarios due to its temporal abstraction, state abstraction, and~sub-task abstraction. These abstractions also helped to reduce the number of Q-values that needed to be stored. The detailed system was presented in~\citep{chan2012}. In~their system, they used three different types of sensory information: a noise-canceling microphone for recognizing human verbal actions, an emWave earclip heart rate sensor for affective arousal level and a webcam for monitoring the activity state (depending upon whether matching card pairs were found or not). They used a two‐stage training process involving offline training followed by online training. The purpose of the first stage was to determine the optimal behaviors for the robot with respect to the card game. The offline training was carried out on a human user simulation model created with the interaction data of ten participants. In the second stage, they aimed to personalize the robot according to the user's state (affective arousal and game state) for different participants in online interactions. The~affective arousal and user activity state formed the user state (e.g., stressed: high arousal and not matching card, pleased: low arousal and matching card). The success of the robot's actions was subject to the improvement of a person's user state from a stressed state to a stress-free state. 
\subsection{Deep Reinforcement Learning}

For natural interaction, it is important that social robots possess human-like social interaction skills, which requires features from high dimensional signals. In these cases, DRL can be useful. In fact, several researchers have begun to examine the applicability of DRL in social robotics~\cite{Qureshi2016, qureshi2017show, Qureshi2018, cuayahuitl2019data, lathuiliere2018, lathuiliere2019, Churamani2018}.

\subsubsection{{Tactile Communication}}
One of the pioneering works using DRL in social robotics was presented by~\cite{Qureshi2016}. Here, a Pepper robot learned to choose among predefined actions for greeting people, based on visual input. In their work, they succeeded to map two different visual input sources, the Pepper robot's RGBD camera and the webcam, to discrete actions (waiting, looking towards the human, hand waving and handshaking) of the robot. The reward was provided by a touch sensor located on the robot's right hand to detect handshaking. The robot received a predefined numerical reward (1 or $-$0.1) based on a successful or unsuccessful handshake. A successful handshake was detected through the external touch sensor. The proposed multimodal DQN consists of two identical streams of CNN for action-value function estimation---one for grayscale frames and another for depth frames. The grayscale and depth images were processed independently, and the Q-values from both streams were fused for selecting the best possible action. This method comprised two phases: the data generation phase and the training phase. In the data generation phase, the Pepper robot interacted with the environment and collected data. After this phase, the training phase began. This two-stage algorithm was useful in that it did not pause the interaction for training.~\citet{Qureshi2016} used 14 days of interaction data where each day of the experiment corresponded to one episode. The same authors applied a variation of DQN, the Multimodal Deep Attention Recurrent Q-Network (MDARQN)~\citep{qureshi2017show}, to the same handshaking scenario in~\citep{Qureshi2016}. In their previous study, the robot was unable to indicate its attention. For adding perceptibility to the robot's actions, a recurrent attention model was used, which enabled the Q-network to focus on certain parts of the input image. Similar to their previous work~\citep{Qureshi2016}, two~identical Q-networks were used (one for grayscale frames and one for depth frames). Each~Q-network consisted of convnets, a Long Short-term Memory (LSTM) network, and an attention network~\citep{xu2015show}. The convnets were used to transform visual frames into feature vectors. The network transforms an input image into D-dimensional $L$ feature vectors, each~of them representing a part of the image $a_t = \{ a_t^1,..., a_t^L \}, a_t^l \in \mathbb{R}^D$. This feature vector was provided as an input to the attention network for generating the annotation vector $z \in \mathbb{R}^D$. The annotation vector $z_t$ is the dynamic representation of a part of an input image at time $t$. $z_t$ is computed with $z_t = \sum_{l = 1}^L \beta_t^l a_t^l$. The LSTM network used the annotation vector $z_t$ for computing the next hidden state. Each of the streams of the MDARQN model were trained by using the back-propagation method. The outputs from the two streams were normalized separately and averaged to create output Q-values of MDARQN. As in their previous work, handshake detection was used for the reward function ($-$0.1 for unsuccessful handshakes and 1 for successful handshakes). The horizontal and vertical axes of the input image were divided into five subregions, and the Q-network enabled to focus on certain parts of the input image. The attention mechanism of the robot used the annotation vector $z_t$ to determine the pixel location to direct maximal attention to the input image. This region selection provided computational benefits by reducing the number of training parameters. Another work from the same authors~\citet{Qureshi2018} proposed an intrinsically motivated DRL approach for the same handshaking scenario. The proposed method utilized three basic events to represent the current state of the interaction, i.e., eye contact, smile, and handshake. These event occurrences were predicted at the next time step according to the state-action pair by a neural network called Pnet. Another neural network called Qnet was employed for action selection policy guided by the intrinsic reward. The~reward was determined based on the prediction error of Pnet, i.e., the error between actual occurrences of events $e(t+1)$ and Pnet's prediction $\hat{e}(t+1)$. An OpenCV-based event detector module provided the labels for three events (i.e., actual event occurrence). The~Qnet was a dual stream deep convolutional neural network mapping pixels to q-values of the actions (wait, look~towards human, wave hand, and shake hand). Pnet was a multi-label classifier which was trained to minimize the prediction error between $\hat{e}$ and $e$ by using the Binary Cross Entropy (BCE) loss function. The reward consisted in predetermined numerical values depending on the prediction error between $e$ and $\hat{e}$. They investigated the impact of three different reward functions named strict, neutral and kind. In all reward functions, if all three events are predicted successfully by Pnet, Qnet receives a reward of 1, if all events are predicted wrong then Qnet gets a reward of $-$0.1. If one or two events are predicted correctly then different reward functions penalize differently, with the strict reward having the highest penalties. The authors reported that the reward functions with more positive reward on incorrect predictions yielded more socially acceptable behavior. They compared the collected total reward from 3 days of experiments in a public place, each day following a different policy (random policy, Qnet policy, and the previously employed method~\citep{Qureshi2016}). The current proposed model led to more human-like behaviors, according to the human~evaluators.

\subsubsection{{No Communication Medium}}
Another study using the Pepper robot and DQN was presented by~\citet{cuayahuitl2019data}. In their scenario, human participants played a ~\qo{Noughts and Crosses} game with two different grids (small and big) against the Pepper robot. They used a CNN for recognizing game moves, i.e., hand-writing on the grid. These visual perceptions and the verbal conversations of the participant were given as an input to their modified DQN. The author modified the Deep Q-Learning with Experience Replay~\citep{mnih2015human} by adding the identification of the worst action set $\hat{A}$. $\hat{A}$ included actions with $min(r(s,a) < 0~\forall a \in A)$ and $A$ is the set of actions leading to win the game. The action selection was done with $\underset{a \in A \backslash \hat{A}}{max} Q(s, a; \theta)$. In other words, the proposed DQN algorithm refines the action set at each step to make the agent learn to infer the effects of its actions (such as selecting the actions that lead to winning or to avoid losing). The reward consisted in predefined numerical values based on the performance of the robot in the game. Therefore, this study does not use any communication medium for reward formulation. The robot received the highest reward in the cases~\qo{about to win} or~\qo{winning}, whereas the robot received the lowest reward in the cases~\qo{about to lose} or~\qo{losing}. 

\subsubsection{{Nonverbal Communication}}
Expressive robot behaviors including facial expressions, gestures, and posture are found to be useful to express the robots' internal states, goals, and desires~\cite{breazeal2009role}. To date, several studies have investigated the production of expressive robot behaviors using DRL, including gaze~\cite{lathuiliere2018,lathuiliere2019} and facial expressions~\cite{Churamani2018}.~\citet{lathuiliere2018} modeled Q-learning with a Long Short Term Memory (LSTM) to fuse audio and visual data for controlling the gaze of the robotic head to direct it towards the targets of interest. The~reward function was defined as $R_t = F_{t+1} + \alpha \sum_{t+1}$ where $\alpha \geq 0$ serves as an adjustment parameter. If the speech sources lie within the camera's field of view, large $\alpha$ values return large rewards, i.e, $\alpha$ permits to give importance to speaking persons. The reward function includes face reward $F_t$ ($\alpha = 0$) and speaker reward ($\alpha > 0$). The number of visible people (face reward) and the presence of speech sources in the camera field of view (speaker reward) were observed from the temporal sequence of camera and microphone observations. The~proposed DRL model was trained on a simulated environment with simulated people moving and speaking, and on the publicly available AVDIAR dataset. In~this offline training, they compared the reward obtained with four different networks: early fusion and late fusion of audio and video data, as well as only audio data and only video data. The authors emphasized the importance of audio-visual fusion in the context of gaze control for HRI. They reported that the proposed method outperformed the handcrafted strategies.~\citet{lathuiliere2019} extended the study presented in~\citep{lathuiliere2018} by investigating the impact of the discount factor, the window size (number of past observations affects the decision), and LSTM network size. They reported that in the experiments with AVDIAR dataset, high discount factors were prone to overfit, whereas in the simulated environment low discount factors resulted in worse performance. Using~smaller window sizes accelerated the training, however, larger window sizes performed better in simulated environment. Changing the LSTM size did not make a substantial difference in the results. In a similar vein,~\citet{Churamani2018} utilized visual and audial data for enabling the Nico robot to express empathy towards the users. They focused on both recognizing the emotions of the user and generating emotions for the robot to display. The~presented model consisted of three modules: an emotion perception module, an intrinsic emotion module, and an emotion expression module. For the perception module, both the visual and audio channels were used to train a Growing-When-Required (GWR) Network. For the emotion expression module, they used a Deep Deterministic Policy Gradient (DDPG) based actor-critic architecture. The reward was the symmetry of the eyebrows and mouth in offline pre-training, whereas in online training the reward was provided by the participant deciding whether the expressed facial expression was appropriate. The Nico robot expressed its emotions through programmable LED displays in the eyebrow and mouth area. 

\subsection{Policy-Based Methods}

\subsubsection{{Higher Level Interaction Dynamics: Comfort}}
In the domain of socially assistive robotics, the robots are expected to be adaptive to their users to some extent, by using social interaction parameters (for example, the interaction distance, the speed of motion and utterances) regarding to the task, to the users' comfort and personality.  Several studies~\cite{mitsunaga2006robot, mitsunaga2008adapting, Tapus2008} examined the Policy Gradient Reinforcement Learning (PGRL) for adapting the robot behaviors using social interaction parameters.~\citet{mitsunaga2006robot, mitsunaga2008adapting} presented a study where the Robovie II robot adjusted its behaviors (i.e., proxemics zones, eye contact ratio, waiting time between utterance and gesture, motion speed) according to comfort and discomfort signals of humans (i.e., body~re-positioning amount and the time spent gazing at the robot). These signals were used as reward. The goal of the robot was to minimize these signals, thereby reducing experienced discomfort in the human interactant. In~\citep{Tapus2008}, an ActiveMedia Pioneer 2-DX mobile robot adapted its personality by changing the interaction distance, speed and frequency of motions, and vocal content (what and how the robot says things). The purpose of this adaptation was to improve the user's task performance. Their reward function was based on user performance, defined as the number of performed exercises. Specifically, the number of performed exercises over the previous 15 s was computed every second and results were averaged over a 60 s period to produce the final evaluation for each policy. They used a threshold for the reward function (7 exercises in the first 10 min) and a time range to adjust the fatigue incurred by the participant. The participant's performance was tracked by the robot through a light-weight motion capture system worn by the participant.

\section{Categorization of RL Approaches in Social Robotics Based on Reward}
\label{sec:categorization}

\begin{figure}[!ht]
\begin{center}
\centering
\captionsetup{justification=centering}
\includegraphics[width=\textwidth]{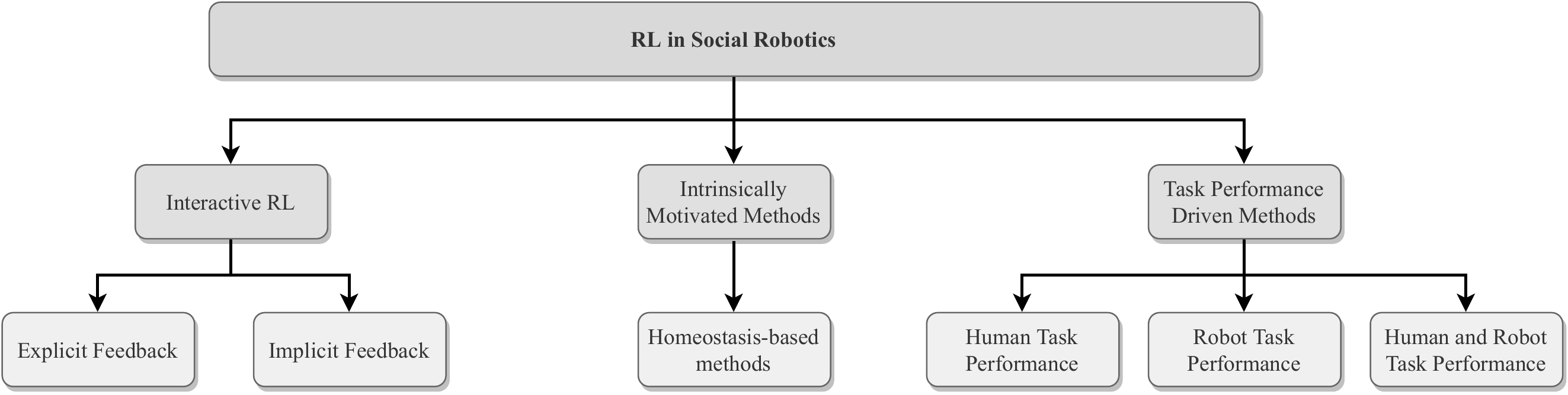}
\caption{Reinforcement Learning approaches in social robotics.}\label{fig:categories}
\end{center}
\end{figure}

We now present a review of the literature but with focus on the reward function. Designing the reward function is perhaps the most crucial step in the implementation of an RL framework. One of the main contributions of this paper is a categorization of different types of reward functions that are used in RL and social robotics. The categorization is given in Figure~\ref{fig:categories}.

As we have already discussed the used RL methods in Section~\ref{categories_rl_based}, they are not included here. Moreover, the evaluation methodologies are also discussed in a separate section \mbox{(see Section~\ref{evaluation_methods}). }

\subsection{Interactive Reinforcement Learning}
\label{interactive_rl}

\begin{figure}[!ht]
\begin{center}
\centering
\captionsetup{justification=centering}
\includegraphics[width=0.65\textwidth]{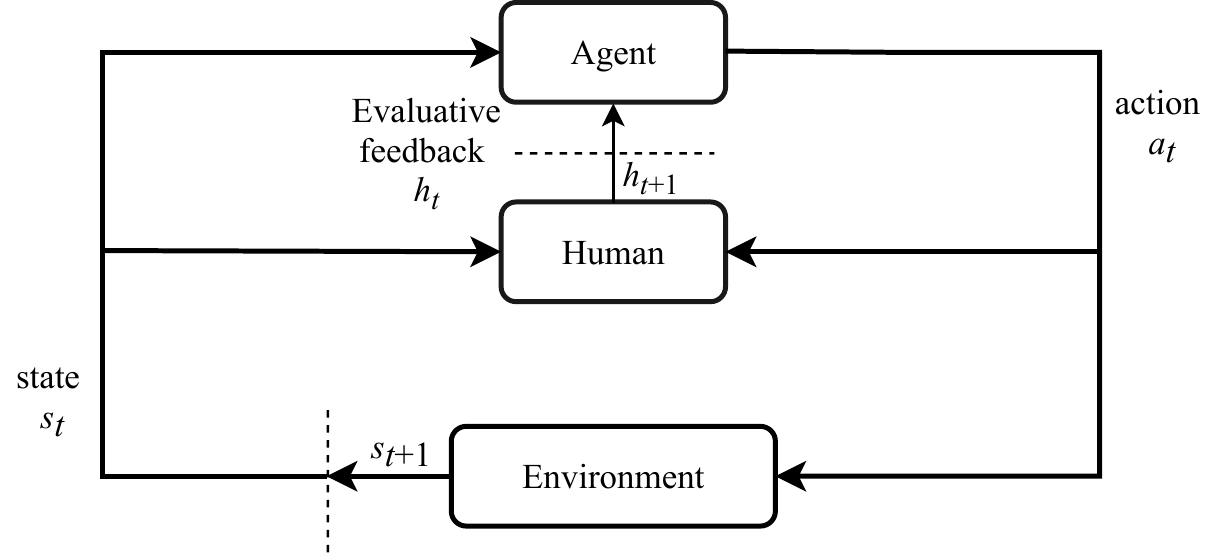}
\caption{Interaction in interactive RL (reproduced from~\citep{li2019human})}\label{fig:interactive_rl}
\end{center}
\end{figure}

Different approaches have been proposed for incorporating the human assistance in the learning process of artificial agents, including learning from human feedback~\cite{knox2012reinforcement, loftin2016learning} and learning from demonstration. Learning from demonstration is beyond the scope of this paper, we focus on learning from human feedback. In traditional RL, the agent receives environmental reward from a predefined reward function. Interactive RL makes use of human feedback in the learning process in combination with or without environmental reward. Interactive RL framework is given in Figure~\ref{fig:interactive_rl}.
Integrating human feedback with RL can be accomplished in different ways, such as via evaluative feedback~\cite{knox2009interactively}, corrective feedback~\cite{celemin2019interactive} or guidance~\cite{thomaz2008teachable}. 

\citet{li2019human} discuss different interpretations of human evaluative feedback in interactive reinforcement learning (referred to as human-centered RL throughout the paper). They distinguish between three types of human evaluative feedback: interactive shaping, learning from categorical feedback and learning from policy feedback. In interactive shaping, human feedback is interpreted as numeric reward, and this reward can be myopic i.e., $\gamma = 0$~\cite{knox2009interactively} or non-myopic i.e., $\gamma$ is different from 0~\cite{knox2015framing}. Human feedback might be erroneous when the task is repetitive. Moreover, human teachers tend to give less frequent feedback (e.g., due to boredom and fatigue) as the learning progresses. Modeling human feedback has been found to be an efficient strategy when the meaning of human-delivered feedback is ambiguous~\cite{loftin2016learning}. ~\citet{loftin2016learning} developed a probabilistic model of human teacher's feedback. They interpret human feedback as categorical feedback, considering that human teachers may have different feedback strategies. In their work, depending on the human teacher's training strategy, a lack of feedback can convey information about the agent's behavior. Human training strategies are categorized into four groups: reward-focused strategy (positive reward for correct actions and no feedback for incorrect actions), punishment-focused strategy (no feedback for correct actions and punishment for incorrect actions), balanced strategy (positive reward for correct actions and punishment for incorrect actions) and inactive strategy (the human teacher rarely provides feedback). Corrective feedback can be categorized under policy feedback. As an example,~\citet{celemin2019interactive} presented a framework named COACH (COrrective Advice Communicated by Humans) which uses human corrective feedback in the action domain as binary signals (i.e., increase or decrease the magnitude of the current action). In their comparison with classical reinforcement learning approaches, they showed that RL agents can benefit from human feedback, i.e., learning progresses faster~\cite{celemin2019interactive}. 
When the agent learns both human feedback and environment reward, the human feedback can be used to guide the agent's exploration~\cite{thomaz2008teachable}. The guidance includes both providing feedback on past actions and guiding the agent in the learning process through future-directed rewards. Human~guidance can reduce the action space by narrowing down the action choices~\cite{thomaz2006reinforcement2}, which~speeds up the training process by accelerating the convergence towards an optimal~policy. 

In the context of HRI, the human can be in the learning loop by way of varying types of inputs, such as providing feedback via a GUI (e.g., by button or mouse clicks). Alternatively, the feedback can be delivered more naturally, via emotions, gestures and speech. Therefore, this category comprises two subcategories: (1) explicit feedback, when~the feedback is direct, provided through an interface such as ratings, and labels; (2) implicit feedback, if the human feedback is spontaneous behavior or reactions such as non-verbal cues and social signals. The terms \qt{explicit feedback} and \qt{implicit feedback} are adopted from~\mbox{\citet{schmidt2000implicit}}'s \qt{implicit interaction} study in human-computer interaction. For a quick summary of the studies, see Table~\ref{tab:irl}. 

\begin{center}
\begin{longtable}{lllll}
\caption{Summary of Interactive Reinforcement Learning approaches in social robotics.} \label{tab:irl} \\

\hline \multicolumn{1}{c}{\textbf{Reference}} & \multicolumn{1}{c}{\textbf{Subcategory}} & 
\multicolumn{1}{c}{\textbf{Type of RL}} &
\multicolumn{1}{c}{\textbf{Reward}} &
\multicolumn{1}{c}{\textbf{Social robot}} \\ \hline 
\endfirsthead

\multicolumn{5}{c}%
{{\bfseries \tablename\ \thetable{} -- continued from previous page}} \\
\hline \multicolumn{1}{c}{\textbf{Reference}} & \multicolumn{1}{c}{\textbf{Subcategory}} & 
\multicolumn{1}{c}{\textbf{Type of RL}} &
\multicolumn{1}{c}{\textbf{Reward}} &
\multicolumn{1}{c}{\textbf{Social robot}} \\ \hline 
\endhead

\hline \multicolumn{5}{r}{{}} \\ 
\endfoot

\hline 
\endlastfoot

\makecell{Barraquand et al. \citep{Barraquand2008}} & \makecell{Explicit \\ feedback} & \makecell{Q-learning} & \makecell{User provides reward by using \\ robot's tactile sensors} & \makecell{Aibo}\\
\midrule 
\makecell{Suay et al. \citep{suay2011effect, suay12}} &  \makecell{Explicit \\ feedback} & \makecell{Q-learning} & \makecell{Human teacher delivers reward \\ or guidance through a GUI} & \makecell{Nao} \\
\midrule 
\makecell{Knox et al. \citep{Knox2013}} & \makecell{Explicit \\ feedback} & \makecell{TAMER} & \makecell{Human teacher provides reward \\ by using a remote} & \makecell{Nexi} \\
\hline 
\makecell{Yang et al. \citep{yang2017}} & \makecell{Explicit \\ feedback} & \makecell{Q-learning} & \makecell{User gives reward by \\ touching robot's tactile sensors} & \makecell{Pepper} \\
\hline 
\makecell{Schneider et al.  \cite{schneider2017exploring}}  &  \makecell{Explicit \\ feedback} & \makecell{Dueling bandit} & \makecell{User provides feedback \\ through a button} & \makecell{Nao} \\
\midrule
\makecell{Churamani et al. \citep{Churamani2018}} & \makecell{Explicit \\ feedback} & \makecell{DDPG} & \makecell{User gives reward whether \\ robot's expression is appropriate to \\ affective context of dialogue} & \makecell{Nico} \\
\midrule
\makecell{Tseng et al. \citep{Tseng18}} & \makecell{Explicit \\ feedback} & \makecell{Modified R-Max} & \makecell{User provides reward \\ through a software} & \makecell{ARIO} \\
\midrule
\makecell{Gamborino et al. \citep{Gamborino18}} & \makecell{Explicit \\ feedback} & \makecell{SARSA} & \makecell{User's transition between \\ bad and good mood state} & \makecell{RoBoHoN} \\
\midrule
\makecell{Ritschel et al. \citep{ritschel2019adaptive}} & \makecell{Explicit \\ feedback} & \makecell{k-armed bandit} & \makecell{User gives reward via buttons} & \makecell{Reeti} \\
\midrule
\makecell{Patompak et al. \citep{patompak2019learning}} & \makecell{Implicit \\ feedback} & \makecell{R-learning} & \makecell{Verbal reward by the user \\ based on robot's social distance} & \makecell{Pepper} \\ 
\midrule
\makecell{Thomaz et al. \citep{Thomaz2007}} & \makecell{Implicit \\ feedback} & \makecell{Q-learning} & \makecell{Human teacher provides \\ reward or guidance} & \makecell{Leonardo} \\
\midrule
\makecell{Thomaz et al. \citep{thomaz08sociallyguided}} & \makecell{Implicit \\ feedback} & \makecell{Q-learning} & \makecell{Human teacher provides guidance\\ through speech or gestures} & \makecell{Leonardo} \\
\midrule
\makecell{Gruneberg et al. \\ \citep{gruneberg2012lesson, gruneberg2013approach}} & \makecell{Implicit \\ feedback} & \makecell{Not specified} & \makecell{Human teacher's smile and frown} & \makecell{Nao} \\ 
\midrule
\makecell{Addo et al. \citep{addo2014applying}} & \makecell{Implicit \\ feedback} & \makecell{Q-learning} & \makecell{Verbal reward of the user} & \makecell{Nao}\\
\midrule
\makecell{Zarinbal et al. \citep{zarinbal2019new}} & \makecell{Implicit \\ feedback} & \makecell{Q-learning} & \makecell{Human teacher's facial expressions} & \makecell{Nao} \\
\midrule
\makecell{Mitsunaga et al. \\ \citep{mitsunaga2006robot, mitsunaga2008adapting}} & \makecell{Implicit \\ feedback} & \makecell{PGRL} & \makecell{Discomfort signals of the user} & \makecell{Robovie II} \\
\midrule
\makecell{Leite et al. \citep{leite2011modelling}} & \makecell{Implicit \\ feedback}&  \makecell{Multi-armed \\ bandit} & \makecell{User's affective cues and \\ task-related features } & \makecell{iCat} \\
\midrule
\makecell{Chiang et al. \citep{Chiang2015}} & \makecell{Implicit \\ feedback} & \makecell{Q-learning} & \makecell{Numerical values based on \\ attention and engagement \\ levels of the user} & \makecell{ARIO} \\
\midrule
\makecell{Gordon et al.  \citep{gordon2016affective}} & \makecell{Implicit \\ feedback} & \makecell{SARSA} & \makecell{Weighted sum of  \\ facial valence and engagement} & \makecell{Tega} \\
\midrule
\makecell{Ritschel et al. \citep{Ritschel2017b}} & \makecell{Implicit \\ feedback} & \makecell{Q-learning} & \makecell{Change in user engagement} & \makecell{Reeti} \\
\midrule
\makecell{Weber et al. \citep{Weber2018a}} & \makecell{Implicit \\ feedback} &  \makecell{Q-learning} & \makecell{Vocal laughter and visual smiles} & \makecell{Reeti} \\
\midrule
\makecell{Park et al. \citep{park2019model}} & \makecell{Implicit \\ feedback} & \makecell{Q-learning} & \makecell{Weighted sum of engagement  \\ and learning} & \makecell{Tega} \\
\makecell{Ramachandran et al. \\ \citep{ramachandran2019}} &  \makecell{Implicit \\ feedback} & \makecell{POMDP} & \makecell{Engagement level of the user} & \makecell{Nao} \\
\midrule
\makecell{Martins et al. \citep{martins2019}} & \makecell{Implicit \\ feedback} & \makecell{Model-based RL \\ and POMDP} & \makecell{The robot's actions' impact \\ on the user} & \makecell{GrowMu} \\
\end{longtable}
\end{center}
 \vspace{-10 mm}

\subsubsection{Explicit Feedback}
\label{explicit_feedback}
In the explicit feedback approach, the feedback of the human teacher is given by direct manipulations and generally through an artificial interface. The human teacher observes the agent's actions and environment states and subsequently provides feedback to the agent through a graphical user interface (GUI) or through the robot's (touch) sensors. In~this approach, the feedback from the human teacher is noiseless and direct in the form of numerical values provided via a button, a Graphical User Interface (GUI), or through the robot's touch sensors. In general, the main purpose of the interaction is to teach the robot to do something in this category. Unlike the explicit feedback category, in the implicit feedback category, the majority of studies include a social scenario such as robot tutoring, robots supporting the human in a game, etc. The studies under this category are~\cite{Barraquand2008,suay2011effect,suay12,Knox2013,yang2017,schneider2017exploring,ritschel2019adaptive}. 

\subsubsection{Implicit Feedback}
\label{implicit_feedback}

Human social signals are widely used as reward in social human-robot interaction. The most commonly used signals are human emotions, as these have a great influence on decision-making~\cite{lerner2015emotion}. Computational models of emotions have been studied by many researchers as part of the agent's decision making architecture, by modelling the RL agents with emotions or incorporating human emotions as an input to the learning process. As an example,~\citet{moerland2018emotion} surveyed RL studies focusing on agent/robot emotions. Since emotions also play an important role in  communication and social robots~\cite{fong2003survey}, there exist various studies considering these aspects for RL and social robotics. 
In the implicit feedback approach, the agent learns from spontaneous natural behavior and reactions of the interactant, i.e., emotions, speech, gestures, etc. This type of feedback is noisy and indirect. In other words, in this approach, human feedback requires pre-processing and the quality of the feedback depends on the perception and recognition algorithms being used. Unlike explicit feedback, the implicit feedback is not provided directly through an interface. Instead, the human's emotions or verbal instructions serve as reward or guidance signals. 
The studies in this category are~\cite{Thomaz2007,thomaz08sociallyguided,gruneberg2012lesson,gruneberg2013approach,patompak2019learning,mitsunaga2006robot,mitsunaga2008adapting,leite2011modelling,addo2014applying,Chiang2015,gordon2016affective,Ritschel2017b,Weber2018a,martins2019,ramachandran2019}. 

\subsection{Methods Using Intrinsic Motivation}
\label{intrinsically_motivated}

It is a common approach to examine the biological and psychological decision-making mechanisms and to use a similar method for autonomous systems. One such approach consists in combining intrinsic motivation with reinforcement learning. Intrinsic motivation is a concept in psychology, which denotes the internal natural drive to explore the environment, as well as gain new knowledge and skills. The activities are performed for inherent satisfaction rather than external rewards~\cite{ryan2000intrinsic}. Researchers have proposed computational approaches that use intrinsic motivation~\cite{oudeyer2009intrinsic}. In intrinsically motivated RL, the main idea is using intrinsic motivations as a form of reward~\cite{chentanez2005intrinsically}. There are different intrinsic motivation models within the RL framework~\cite{oudeyer2008can}. However, in social robotics, the~idea of maintaining the internal needs of the robot (detailed in Section \ref{homeostasis}) has received much attention~\cite{Silva12, perula2019bioinspired, maroto2018bio, castro2014learning, castro2013autonomous, Castro-Gonzalez2011, malfaz2011biologically}. One exception is~\cite{Qureshi2018}, in which prediction error of social event occurrences was used as intrinsic motivation. For a quick summary, see Table~\ref{tab:intrinsic}. 

\begin{table}[!ht]
\caption{Summary of Intrinsically Motivated Methods in social robotics.} \label{tab:intrinsic} 

\setlength{\tabcolsep}{2.4mm}\begin{tabular}{ccccc}
\toprule \multicolumn{1}{c}{\textbf{Reference}} & \multicolumn{1}{c}{\textbf{Subcategory}} & 
\multicolumn{1}{c}{\textbf{Type of RL}} &
\multicolumn{1}{c}{\textbf{Reward}} &
\multicolumn{1}{c}{\textbf{Social Robot}} \\ 

\midrule
\makecell{Malfaz et al. \cite{malfaz2011biologically}} &  \makecell{Homeostasis based} & \makecell{Q-learning} & \makecell{Wellbeing of the robot} & \makecell{Maggie} \\
\midrule
\makecell{Castro-Gonzalez et al. \\ \cite{Castro-Gonzalez2011, castro2013autonomous, castro2014learning}} & \makecell{Homeostasis \\ based} & \makecell{Object \\ Q-learning} & \makecell{Variation of robot's wellbeing} & \makecell{Maggie} \\
\midrule
\makecell{Maroto et al.  \cite{maroto2018bio}} & \makecell{Homeostasis \\ based} & \makecell{Q-learning} & \makecell{Maximization of robot's well-being} &\makecell{ Mini}\\
\midrule
\makecell{Perula et al. \cite{perula2019bioinspired}} & \makecell{Homeostasis \\ based} & \makecell{Q-learning} & \makecell{Well-being of the robot} & \makecell{Mini}\\ 
\midrule
\makecell{Da Silva et al. \cite{Silva12}} & \makecell{-} & \makecell{Economic TG} & \makecell{Generated on the basis of \\ internal state estimate} & \makecell{Robotic head} \\
\midrule
\makecell{Qureshi et al. \cite{Qureshi2018}} & \makecell{-} & \makecell{DQN} & \makecell{Prediction error of an action \\ conditional prediction network} & \makecell{Pepper} \\
\bottomrule
\end{tabular}
\end{table}

\subsubsection*{Homeostasis-Based Methods}
\label{homeostasis}

Homeostasis, as defined by~\citet{cannon1939wisdom}, refers to a continuous process of maintaining an optimal internal state in the physiological condition of the body for survival. \citet{berridge2004motivation} explains homeostasis motivation with a thermostat example that behaves as a regulatory system by continuously measuring the actual room temperature and comparing it with a predefined set point, and activating the air conditioning system if the measured temperature deviates from the predefined set point. In the same manner, the body maintains its internal equilibrium through a variety of voluntary and involuntary processes and behaviors. The homeostasis-based RL in social robotics is presented in~\cite{perula2019bioinspired, maroto2018bio, castro2014learning, castro2013autonomous, Castro-Gonzalez2011, malfaz2011biologically}. These studies introduced a biologically inspired approach that depends on homeostasis. The robot's goal was to keep its well-being as high as possible while considering both internal and external circumstances. The common theme in these studies is that the robot has motivations and drives (needs), where each drive has a connection with a motivation as in Equation (\ref{motivation_drive}). 
\begin{equation}
\label{motivation_drive}
\begin{split}
    & {if}~~D_{i} < L_{d}~~{then}~~M_{i} = 0 \\
    & {if}~~D_{i}\geq L_{d}~~{then}~~M_{i} = D_{i}+w_{i}
\end{split}
\end{equation}

Motivations whose drives are below the activation levels do not initiate a robot behavior. This was formulated as $if~~D_{i} < L_{d}~~then~~M_{i} = 0$ where $D_{i}$ is a drive, $L_{d}$~the activation level, and $M_{i}$ is the related motivation. The motivation depends on two factors: the associated drive and the presence of an external stimulus, this was formulated as $if~~D_{i}\geq L_{d}~~then~~M_{i} = D_{i}+w_{i}$ where $w_{i}$ is the related external stimulus. These motivations serve as action stimulation to satiate the drives. A drive can be seen as a deficit that leads the agent to take action in order to alleviate this deficit and maintain an internal equilibrium. The ideal value for a drive is zero, corresponding to the absence of need. The robot learns how to act in order to maintain its drives within an acceptable range, i.e., to maintain its well-being. The well-being of the robot was defined as: \begin{equation}
    Wb = Wb_{ideal} - \sum_{i} \alpha_{i} D_{i}
\end{equation}
where $Wb_{ideal}$ is the value of the well-being when all drives are satiated, and $\alpha_{i}$ is the set of the personality factors that weight the importance of each drive. The variation of the robot's well-being is used as reward signal and calculated with the Equation (\ref{wellbeing_variation})
\begin{equation}
    \label{wellbeing_variation}
    \Delta Wb = Wb_{t} - Wb_{t-1}
\end{equation}
 i.e., the difference between the current well-being $Wb_t$ and the well-being in the previous step $Wb_{t-1}$.
 
In several works~\cite{castro2014learning, castro2013autonomous, Castro-Gonzalez2011}, a variation of the traditional Q-learning algorithm was used in addition to the homeostasis-based approach. In all of these, the authors referred to the proposed algorithm as Object Q-learning. In this approach, there are actions associated with each object in the environment, and the robot considers its state in relation to every object independently. Thus, there is an assumption that an action execution in relation to a certain object does not influence the state of the robot in relation to other objects. However, in reality, an action execution may create collateral effects. In other words, an~action associated with a particular object, e.g., approaching it, may affect the robot's state in relation to other objects, e.g., moving away from them. The update of Q-values accounted for these collateral effects. The purpose of this simplification was to reduce the number of states during the learning process. In their experiments, to reduce the state space, the robot learned what to do with each object without considering its relation to other objects. In other words, they assumed that an action execution associated with a certain object will not affect the state of the robot in relation to the rest of the objects. The proposed algorithm was implemented on the social robot Maggie that lived in a laboratory and interacted with several objects in the environment (e.g., a music player, a docking station, or humans).~\citet{castro2013autonomous} appears to be closely linked to the other papers discussed here with one difference being that a discrete emotion, fear, was used as one of the motivations. Unlike other motivation-drive pairs, no drive was associated with the \qo{fear motivation} (i.e., fear is not a deficiency of any need). \qo{Fear motivation was} linked to dangerous situations (that can cause damage the robot) and directed the robot to a secure state. As an example, the motivation \qo{social} was not updated if the user who occasionally hit the robot was around.

\subsection{Methods Driven by Task Performance}
\label{task_performance}
Task performance denotes the effectiveness with which an agent performs a given task, and the performance metrics can vary for different tasks. In these methods, the design of the reward function is based on task-driven measures, which often include some problem-specific information, especially the task performance of the robot, task performance of the human, or both. For a quick summary, see~Table~\ref{tab:task_perf}. 

\begin{table}[!ht]
\caption{Summary of Task performance driven methods in social robotics.} \label{tab:task_perf}

\setlength{\tabcolsep}{2.3mm}\begin{tabular}{ccccc}
\toprule \multicolumn{1}{c}{\textbf{Reference}} & \multicolumn{1}{c}{\textbf{Subcategory}} & 
\multicolumn{1}{c}{\textbf{Type of RL}} &
\multicolumn{1}{c}{\textbf{Reward}} &
\multicolumn{1}{c}{\textbf{Social Robot}} \\ \midrule

\makecell{Tapus et al. \cite{Tapus2008}} & \makecell{Human task \\ performance} & \makecell{PGRL} & \makecell{User performance} & \makecell{Pioneer 2-DX} \\
\midrule
\makecell{Gao et al. \cite{gao2018robot}} & \makecell{Human task \\ performance} & \makecell{Multi-arm \\ bandit} & \makecell{User task performance and \\ user's verbal feedback} & \makecell{Pepper} \\
\midrule
\makecell{Chan et al.  \cite{chan2011learning,chan2011minimizing}} &  \makecell{Human and robot \\ task performance } & \makecell{MAXQ} & \makecell{Success of the robot's actions \\ in helping or improving user's \\ affect and task performance} & \makecell{Brian 2.0} \\
\midrule
\makecell{Chan et al. \cite{chan2012}} & \makecell{Human and robot \\ task performance} & \makecell{MAXQ} & \makecell{Task performance of \\ human and robot} & \makecell{Brian 2.0} \\
\midrule

\makecell{Moro et al. \cite{moro18}} & \makecell{Human and robot \\ task performance} & \makecell{Q-learning} & \makecell{Numerical numbers based on \\ robot's performance on user's \\ activity state} & \makecell{Casper} \\
\midrule
\makecell{Nejat et al. \cite{nejat2008can}} & \makecell{Robot task \\ performance} & \makecell{Q-learning} & \makecell{User provides verbal feedback} & \makecell{Brian} \\
\midrule
\makecell{Ranatunga et al. \cite{Ranatunga2011}} & \makecell{Robot task \\ performance} & \makecell{TD($\lambda$)} & \makecell{Head and eye kinematic \\ scheme of the robot} & \makecell{Zeno} \\
\midrule
\makecell{Keizer et al. \cite{Keizer2014}} & \makecell{Robot task \\ performance} & \makecell{Monte-Carlo \\ control} & \makecell{The robot's performance \\ as a bartender} & \makecell{iCat} \\
\midrule
\makecell{Qureshi et al. \cite{Qureshi2016}} & \makecell{Robot task \\ performance} & \makecell{Multimodal \\ DQN} & \makecell{Numerical values based on \\ robot's handshake success} & \makecell{Pepper} \\
\midrule

\makecell{Papaioannou et al. \\ \cite{papaioannou2017hybrid}} & \makecell{Robot task \\ performance} & \makecell{Q-learning} & \makecell{Task completion of the robot} & \makecell{Pepper} \\
\midrule

\makecell{Qureshi et al. \cite{qureshi2017show}} & \makecell{Robot task \\ performance} & \makecell{MDARQN} & \makecell{Numerical values based on \\ robot's handshaking success} & \makecell{Pepper} \\
\midrule
\makecell{Hemminghaus et al. \cite{hemminghaus2017}} & \makecell{Robot task \\ performance}  & \makecell{Q-learning} & \makecell{Robot's task performance and \\ execution cost of the robot's action} & \makecell{Furhat} \\
\midrule
\makecell{Chen et al. \cite{Chen2018}} & \makecell{Robot task \\ performance} & \makecell{Q-learning} & \makecell{Numerical values based on \\ correctly completed tasks} & \makecell{Mobile robots} 
\\
\midrule
\makecell{Ritschel et al.  \cite{ritschel2018drink}} & \makecell{Robot task \\ performance} & \makecell{n-armed \\ bandit} & \makecell{Robot's performance at convincing \\ user to select healthy drink} & \makecell{Reeti} \\
\midrule
\makecell{Lathuiliere et al. \cite{lathuiliere2019, lathuiliere2018}} & \makecell{Robot task \\ performance} & \makecell{DQN} & \makecell{Number of observed faces and \\ presence of speech sources \\ in the visual field} & \makecell{Nao} \\
\midrule
\makecell{Cuayahuitl  \cite{cuayahuitl2019data}} & \makecell{Robot task \\ performance} & \makecell{DQN} & \makecell{Numerical values based on \\robot's performance in the game} & \makecell{Pepper} \\
\bottomrule
\end{tabular}
\end{table}

\subsubsection{Human Task Performance Driven Methods}

In these human task performance driven methods, the reward function is based on the user's success in performing a task related to the interaction with the robot. The studies in this category are~\cite{Tapus2008,gao2018robot}. 

\subsubsection{Robot Task Performance Driven Methods}

In these methods, the reward design depends on the robot's task performance. Robot~behaviors that satisfy the user's preferences, accurate completion of the task, finishing the task within a desired amount of time, visiting certain states, and robot actions that benefit or satisfy the user are examples for task performance measures. The studies in this category are~\cite{nejat2008can, Ranatunga2011, Keizer2014, Qureshi2016, papaioannou2017hybrid, hemminghaus2017, lathuiliere2018, Chen2018, lathuiliere2019, ritschel2018drink, cuayahuitl2019data}. 

\subsubsection{Human and Robot Task Performance Driven Methods}

In the previous two sections, we listed the studies using task performance of the robot and human as reward signal. There are also studies that use a combination of the human's and the robot's task performance as reward signal. As an example, in~\cite{chan2012,chan2011learning} the robot received the highest reward if the user completed the task successfully. The robot also received reward for its actions that were suitable for the current situation. Likewise, in~\cite{moro18}, the robot was rewarded based on actions that transitioned the user into a desirable state (e.g., completing the activity). Other papers in this category are~\cite{chan2012, chan2011minimizing, chan2011learning}.

\section{Evaluation Methodologies}
\label{evaluation_methods}

The past decade has seen a rapid growth of social robotics in diverse uncontrolled environments such as homes, schools, hospitals, shopping centers, or museums. In this review, we have seen various application domains in a range of fields including therapy~\cite{hemminghaus2017}, eldercare~\cite{nejat2008can}, entertainment~\cite{Weber2018a}, navigation~\cite{patompak2019learning}, healthcare~\cite{schneider2017exploring}, education~\cite{park2019model}, personal~robots~\cite{maroto2018bio}, and rehabilitation~\cite{Tapus2008}. Research in the field of social robotics and human-robot interaction becomes crucial as more and more robots are entering our lives. This brings many challenges as social robots are required to deal with dynamic and stochastic elements in social interaction in addition to the challenges in robotics. Besides these challenges, validation of social robotics systems with users necessitates efficient evaluation methodologies. Recent studies underline the importance of evaluation and assessment methodologies in HRI~\cite{sim2015extensive}. However, developing a standardized evaluation procedure still remains a difficult task. Furthermore, in RL-based robotic systems, there is a need to explore various human-level factors (personal preferences, attitudes, emotions, etc.) to assure that the learned policy leads to better HRI. Additionally, how can we evaluate whether the learned policy conveys the intended social skill(s)? As an example, in~\citet{Qureshi2016, qureshi2017show, Qureshi2018}'s study, the~model performance on a test dataset was evaluated by three volunteers who judged if the robot's action was an appropriate one for the current scenario. In~\cite{Churamani2018}, there~both annotators and participants rated whether the robot was able to associate the facial expressions with the conversation context. The independent annotators' ratings were higher than the participants', which, as the authors argued, might be explained by discrepancies between the participants' actual expressed emotion and the intended emotion. In~such cases, additional sensory information could be useful for validating that the adaptive robot behaviors lead to better HRI. For example,~\citet{park2019model} analyzed the body poses and electrodermal activity (EDA) of the participants to check their correlation with participant's engagement. This~kind of approach could be used to support subjective evaluations. A comparative evaluation methodology considering both the learned policy and the user's experience about the interaction is another way of evaluation. As~an example~\cite{chan2012, mitsunaga2006robot, mitsunaga2008adapting, Chiang2015, patompak2019learning} presented the policy for each participant as well as a discussion on the effectiveness of the robot behavior on the user based on their comments and subjective~evaluations.

The papers in the scope of this manuscript used different evaluation and assessment methodologies for their algorithms and for their systems with users. We identify three types of evaluation methodologies: (1) an evaluation from the algorithm point of view, (2) evaluation and assessment of user experience-related subjective measures, and (3)~evaluation of both learning algorithm-related factors and user experience-related factors. Several~studies only reported the self-rated questionnaire results~\cite{leite2011modelling} or user opinions~\cite{addo2014applying}.
There are also studies which do not include any evaluation, and only a short discussion regarding the learned policy~\cite{yang2017, gruneberg2012lesson, gruneberg2013approach, Ritschel2017b}

The cumulative collected reward over time is the most commonly used evaluation method. As learning progresses, the frequency of negative rewards is expected to decrease and positive rewards are expected to increase. Thus, the cumulative reward and comparing the reward across different settings and variations of algorithms are one of the measures for evaluating the efficiency of learning~\cite{lathuiliere2019, Tseng18, Barraquand2008, lathuiliere2018, martins2019}. The evolution of the learning algorithm over time (e.g., the evolution of $Q$ 
values) is another evaluation method. Several studies presented only the learning evolution of their system without mentioning how a participant would perceive the learned robot behaviors~\cite{maroto2018bio, castro2014learning, moro18, Silva12, Castro-Gonzalez2011, malfaz2011biologically, castro2013autonomous, perula2019bioinspired}. Comparison of user experiences (e.g., learning gains of children) for adaptive and non-adaptive robot is another way of evaluation~\cite{ramachandran2019, gordon2016affective}. We also see evaluation by using only interaction related objective measures such as the frequency of turn-taking and dialogue duration with the robot~\cite{papaioannou2017hybrid}.
Task-related evaluation measures (i.g., the number of moves needed to solve a game with an adaptive versus a random robot) together with Q-matrix~\cite{hemminghaus2017}, or average task success and average reward~\cite{cuayahuitl2019data} are used. In some IRL studies, the purpose is only teaching a robot. In these studies, evaluation metrics are training time~\cite{Knox2013, suay2011effect}, or training related parameters (e.g., the amount of positive and negative feedback)~\cite{thomaz08sociallyguided}.

Studies reporting both subjective user opinions and algorithm related measures are~\mbox{\cite{suay12, Weber2018a, Tapus2008, schneider2017exploring, ritschel2018drink}}. Interaction related objective measures such as interaction duration, distance to the robot, preferred motion speed of the robot in combination with questionnaires are other measures for evaluating the efficiency of the learned policy. Studies also use a comparison of different algorithms in terms of average steps, average reward, average execution time together with questionnaires~\cite{Chen2018}, and the number of times the preferences of the trainer match with the agent's action~\cite{Gamborino18}, reward and discussion of observations from the experiments~\cite{ritschel2018drink}, questionnaires and task-related parameters (e.g., time to complete the task)~\cite{gao2018robot}.

\section{Discussion}
\label{discussions}

In this paper, we present the RL approaches in social robotics. In virtual game environments (e.g., Atari, Go, etc.) which are commonly used testbeds for RL implementations, the~goal is well defined (e.g., getting higher scores, accomplishing a game level, or winning the game). In social robotics, the goal is not that clear. 
\mbox{Still, we argue} that social robots could provide a unique potential testbed for RL implementations in real-world scenarios, in a sense that they can deal with transparency issues by showing their internal states through social cues (e.g., facial expressions, gaze, speech, LEDs on their body, tablet). In Section~\ref{sec:categorization}, we presented RL approaches based on reward types. IRL with implicit reward is the most widely used approach in social robotics since human social cues occur naturally during the interaction. However, the~change in social cues can be slow, which leads to sparse reward. A combination of the reward approaches presented Section~\ref{sec:categorization}, namely intrinsically motivated methods, IRL with implicit feedback, and task performance-driven methods could be an approach to deal with the sparse reward problem. This way the robot could receive a reward even when there is no dramatic change in social cues or the task is not completed in one step. Similar to the homeostasis-based approaches, combining emotional models for robots' decision-making mechanisms could be helpful. The interested reader may refer to~\cite{moerland2018emotion} which presents a thorough analysis and discussion of computational emotional models incorporated within RL. Th sparse reward problem is not the only problem in real-world social HRI. We continue to the discussion with the proposed solutions for real-world RL problems in Section~\ref{section:real_world_rl}. Later on, we present possible interesting future directions in Section~\ref{future_outlook}.

\subsection{Proposed Solutions to Real-World RL Problems}
\label{section:real_world_rl}
RL is a powerful and versatile algorithmic tool and has been shown to perform better than humans in simulated environments~\cite{mnih2015human} 
However, the progress on applying RL methods to real-world systems is not so advanced yet. This is due to the complexity of the real-world.~\citet{dulac2019challenges} discuss nine challenges of realizing RL on real-world systems. Here, we discuss these challenges and how some papers tackled them in real-world HRI with social robots. 

The first mentioned challenge is~\qt{training off-line from the fixed logs of an external behavior policy}. 
This challenge applies to HRI since users would not tolerate the long pauses and action delays of the social robot. As an example,~\citet{Qureshi2016} suggested an approach where they divided training into two stages. In the first stage, the robot interacts with the environment and gathers data, whereas in the second stage the robot rests and trains. 

The second challenge is~\qt{learning on the real system from limited samples}. This~challenge is especially valid for HRI since the interaction time with the users is limited in controlled lab experiments. Moreover, users get bored and tired with longer interaction duration. As mentioned~\cite{dulac2019challenges} exploration must be limited. As an example, in~\cite{maroto2018bio, perula2019bioinspired}, exploration and exploitation phases are separated. A predefined duration is set for the exploration phase, in which the robot runs through all possible states and actions. Moreover, they also decreased the learning rate $\alpha$ throughout the exploration phase to increase the importance of previously learned information as the learning progresses. In~the exploitation phase, they set $\alpha$ to 0. As mentioned in~\cite{dulac2019challenges}, for improving the sample efficiency expert demonstrations can be beneficial to avoid learning from scratch. For~example,~\mbox{\citet{moro18}} combined LfD with Q-learning for a Casper robot helping older people in a tea making scenario. Another mentioned solution was model-based RL, of which we~see two examples in social robotics~\cite{Tseng18, martins2019}. In addition, long-term interactions (several sessions~\cite{gordon2016affective, ramachandran2019, park2019model}) are important for HRI and could be beneficial for RL to collect~samples.

The third challenge is~\qt{high-dimensional continuous state and action spaces}. 
In the context of social robotics, the problem also needs to be simplified due to the low onboard computational power of most platforms. That might be another reason for a small set of actions in the reviewed papers. To overcome this challenge we see several approaches. As an example, human guidance was found to effective~\cite{suay2011effect}, as well as Object Q-learning~\mbox{\cite{castro2014learning, castro2013autonomous, Castro-Gonzalez2011}} and action elimination~\cite{cuayahuitl2019data}. 

The fourth challenge is~\qt{safety constraints that should never or at least rarely be violated}. The mentioned approaches for this challenge in~\cite{dulac2019challenges} include imposing safety constraints during the training. In the current literature, social robot interactions are generally conducted in a controlled laboratory environment and the researchers are available to intervene if any problems occur. Therefore, this challenge seems to get little attention. 

The fifth challenge is~\qt{tasks that may be partially observable, alternatively viewed as non-stationary or stochastic}. We see several attempts in social robotics to deal with this challenge such as in POMDP based approaches~\cite{martins2019, ramachandran2019}, and in DRL where several frames are stacked together for incorporating the history of the agent observations. Another mentioned approach to deal with this challenge was using recurrent networks which were applied in~\cite{Silva12}. 

The sixth challenge is~\qt{reward functions that are unspecified, multi-objective, or risk-sensitive}. Some papers that use simulated environments for training and testing on real-world interactions. In these papers, there are different reward functions for the simulated world and the real-world. Generally, the real-world reward functions are simplified to one parameter such as feedback of the user or predefined numeric numbers, whereas the simulated world reward functions are more complex including several parameters. 

The~seventh challenge is~\qt{system operators who desire explainable policies and actions}. This~is particularly valid for social robotics, since ambiguous robot behaviors might affect the user's willingness to interact again. Moreover, if the human trains the robot, the intention and internal state of the robot becomes crucial for the success of the training. As an example,~\citet{Knox2013} discussed the transparency challenges and their effect on the training time.~\citet{thomaz08sociallyguided} observed that participants had a tendency to wait for eye contact with the robot before saying the next utterance while training the robot. These kinds of social cues on the robot could be used for explaining its actions and internal states. 

The eighth challenge is~\qt{inference that must happen in real-time at the control frequency of the system}. The real-world is slower than the simulated world both in reaction and data generation. To deal with this challenge, several researchers used an additional interface between the robot and the human, so that the inference is received from the interface rather than robot~control. 

The ninth challenge is~\qt{large and/or unknown delays in the system actuators, sensors, or rewards}. We see several approaches to deal with this challenge, as an example~\cite{Barraquand2008} considered to increase the effect of human-delivered positive reward in larger time frames and to decrease the effect of negative reward in a shorter time frame. Another approach was estimating reward from natural human feedback using the gamma distribution~\cite{Tseng18}. 

\subsection{Future Outlook}
\label{future_outlook}

There are still many interesting potential problems and open questions to be solved in RL for social robotics. Applications on physically embodied robots are limited due to the enormous challenge of complexity and uncertainty in real-world social interactions. The increased prevalence of RL in physical social robots will shed further light on this topic. Another unanswered question is how RL-based social robotics may include the generation of reward signals from ambiguous or conflicting sources of implicit feedback, and how learned skills can be transferred to different robots. Further work could also investigate larger state-action spaces, as current studies are mostly limited to  a small sets. 

\textls[-20]{Despite the fact that there are goal-oriented approaches for social robot learning~\mbox{\cite{lockerd2004tutelage, liu2014interactive}},} in the current literature, the social robot that learns through RL has only one goal, such~as performing a single task and optimizing a single reward function. However, in many real-world scenarios, a robot may need to perform a diverse set of tasks. As an example, socially assistive robots designed with the purpose of assisting older people in their houses may need to accomplish several tasks such as medication reminders, detecting issues, informing caregivers, and managing plans. Multi-goal RL enables an agent to learn multiple goals, hence the agent can generalize the desired behavior and transfer skills to unseen goals and tasks~\cite{bai2019guided}. This has been applied on robotic manipulation tasks in a simulated environment~\cite{bai2019guided}. However, applying the multi-goal RL framework to social robots would be a fruitful area for future work. 

Another interesting future direction might be the application of multi-objective RL in social robotics. The task efficiency and user satisfaction can be two objectives where the robot would try to maximize both objectives by formalizing the problem as a multi-objective MDP.
As an example,~\citet{hao2019emotion} presents a multi-objective weighted RL in which the agent had two objectives: minimizing the cost of service execution and eliminating the user's negative emotions. We refer the interested reader to the survey on multi-objective decision making for a more detailed explanation of the topic~\cite{roijers2013survey}.

Recent developments in the field of deep neural networks have led to an increasing interest in DRL. Applying DRL in social robotics has also received recent attention, \mbox{however}, studies focused on small sets of actions and single task scenarios. In this regard, social~robots with larger sets of actions would be a promising area for further work. \mbox{Another} future direction can be a further investigation of hyper-parameters of RL in social robotics. This was briefly discussed in~\cite{Keizer2014}, as an example, in turn-based interactions relatively small discount factors (i.e., $0.7\leq\gamma\leq0.95$) are more common, whereas for the frame-based interactions with rather long trajectories, higher discount factors seem to be more suitable (i.e., $\gamma\geq0.99$). In deep networks, the selection of different hyper-parameters affects the accuracy of the algorithm~\cite{zhang2017intent}. This also applies to DRL,~\citet{lathuiliere2019} presented several experiments to evaluate the impact of some of the principal parameters of their deep network structure.

Thus far, model-free RL learning a value function or a policy through trial and error is the most commonly used approach in social robotics. However, model-based RL that focuses on learning a transition model of the environment serving as a simulation remains to be further explored. In particular, having a user model can ease the learning process. Although it is difficult to model human reactions, having a model can play a crucial role in reducing the number of required interactions in the real-world. The~model-based approach can also help with the problem of hardware depreciation which may arise in model-free RL in robotics because of the considerable amount of interaction time. Simulating the interaction environment can ease the training without manual interventions and a need for maintenance. Nonetheless, transferring the learned policies in simulation directly to the physical robot may not be trivial due to undermodeling and uncertainty about system dynamics~\cite{kober2013reinforcement}. A common limitation is that most of the works are not generalizable, i.e., utilizing the knowledge learned by one robot on the other or utilizing the task knowledge for other tasks. The Google AI team trained a model-based Deep Planning Network (PlaNet) agent which achieved six different tasks (i.e.,  cartpole swing-up, cartpole~balance, finger spin, cheetah run, etc.)~\cite{hafner2018learning}. A similar approach for a physical social robot would be an interesting future direction. 

RL problems are formalized as MDPs in fully observable environments. However, in the case of HRI not all of the required observations are available, due to the underlying effect of psychological states on human behavior. It has been demonstrated that POMDPs are able to model the uncertainties and inherent interaction ambiguities in real-world HRI scenarios~\cite{kostavelis2017pomdp}.~\citet{hausknecht2015deep} proposed a method that couples a Long Short Term Memory with a Deep Q-Network to handle the noisy observations characteristic of POMDPs. A~similar approach would be useful in social robotics problems to better capture the dynamics of the environment. We included two examples of POMDP approaches in social robotics,~\cite{ramachandran2019,martins2019}. Further investigation would constitute an interesting line of~research.

\section{Conclusion}
\label{conclusions}
In this work, we give an overview of the work on RL in social robotics. We surveyed the literature and presented a thorough analysis of RL approaches in social robotics. \textls[20]{{\mbox{Social}~robots have two important characteristics: physical embodiment and interaction/}communication} capabilities. Therefore, we included studies with physically embodied robots. Moreover, we categorize the papers based on the used RL type. In this categorization, we discuss and group the papers based on the communication medium used for reward formulation. Considering the importance of designing the reward function, we also categorize the papers based on the nature of the reward. The evaluation methods of the papers are also grouped by whether or not they use subjective and algorithmic metrics. We~then provide a discussion in the view of real-world RL challenges and proposed solutions. The points that remain to be explored, including the approaches that have thus far received less attention are also given in the discussion section. To conclude, despite tremendous leaps in computing power and advances in learning methods, we are still a long way from general-purpose, robust, and versatile social robots that can learn several skills from naive users with real-world interactions. In spite of the immediate challenges, we see steady progress of RL applications in social robotics with an increasing interest in recent years. 

\vspace{6pt}

\section*{Author Contributions}
N.A. was the main author responsible for conducting literature research, methodology definition, and paper writing. A.L. supervised the study, and has been involved in structuring and writing the paper. All authors have read and agreed to the published version of the manuscript.

\section*{Funding}
This research was funded by European Union's Horizon 2020 research and innovation program under the Marie Sk\l{}odowska-Curie grant agreement No 721619 for the SOCRATES project.

\bibliographystyle{ACM-Reference-Format}

\end{document}